%% file: main.tex
\documentclass{article} 
\usepackage{iclr2023_conference,times}

\usepackage{ amsmath }
\usepackage{ subcaption }
\usepackage{ math_commands }
\usepackage{ other_commands }
\usepackage{ hyperref }
\usepackage{ url }
\usepackage{ array }
\usepackage[algoruled]{ algorithm2e }
\usepackage{ wrapfig }
\usepackage[subtle]{ savetrees }

\renewcommand{\eqref}[1]{Equation~\ref{#1}}

\title{Bitrate-Constrained DRO: Beyond Worst \\ Case Robustness To Unknown Group Shifts}

\author{%
Amrith Setlur$^{1}$ \phantom{\thanks{Correspondence can be sent to \href{mailto:asetlur@cs.cmu.edu}
{asetlur@cs.cmu.edu}.}} \quad Don Dennis$^{1}$  \quad Benjamin Eysenbach$^{1}$ \\ \textbf{Aditi Raghunathan}$^{1}$ \quad \textbf{Chelsea Finn}$^{2}$ \quad \textbf{Virginia Smith}$^{1}$ \quad \textbf{Sergey Levine}$^{3}$ \\[5pt]
     $^1$ Carnegie Mellon University \quad $^2$ Stanford University \quad $^3$ UC Berkeley \\
}

\definecolor{carnelian}{rgb}{0.7, 0.11, 0.11}

\iclrfinalcopy 
\begin{document}

\maketitle

\vspace{-.1in}
\begin{abstract}

Training machine learning models robust to distribution shifts 
is critical for real-world applications.
Some robust training algorithms (\eg Group DRO) specialize to group shifts and require group information on all training points.
Other methods (\eg CVaR DRO) that do not need group annotations can be overly conservative, since they naively upweight high loss points which may form a contrived set that does not correspond to any meaningful group in the real world (\eg when the high loss points are randomly mislabeled training points).   
In this work, we address limitations in prior approaches by assuming a more nuanced form of group shift: conditioned on the label, we assume that the true group function (indicator over group) is simple. 
For example, we may expect that group shifts occur along low bitrate features (\eg  image background, lighting).
Thus, we aim to learn a model that maintains high accuracy on simple group functions realized by these low bitrate features, that need not spend valuable model capacity achieving high accuracy on contrived groups of examples.
Based on this, we consider the two-player game formulation of DRO where the adversary's capacity is bitrate-constrained.
Our resulting practical algorithm, Bitrate-Constrained DRO (\bdro), does not require group information on training samples yet matches the performance of Group DRO on datasets that have training group annotations and that of CVaR DRO on long-tailed distributions.
Our theoretical analysis reveals that in some settings \bdro objective can provably yield statistically efficient and less conservative solutions than unconstrained CVaR DRO.

\end{abstract}

\input{sections/introduction}
\input{sections/relwork}
\input{sections/preliminaries}

\input{sections/bDRO}
\input{sections/theory}

\input{sections/experiments}

\input{sections/conclusion}

\bibliography{iclr2023_conference}
\bibliographystyle{iclr2023_conference}
\clearpage
\appendix
\input{sections/appendix}

\end{document}

%% file: sections/introduction.tex
\vspace{-0.5em}
\section{Introduction}
\label{sec:introduction}
\vspace{-0.5em}

Machine learning models may perform poorly when tested on distributions that differ from the training distribution.
A common form of distribution shift is \emph{group} shift, where the source and target differ only in the marginal distribution over finite groups or sub-populations, with no change in group conditionals~\citep{oren2019distributionally,duchi2019distributionally} (\eg when the groups are defined by spurious correlations and the target distribution upsamples the group where the correlation is absent~\cite{sagawa2019distributionally}). 

\begin{figure}[!t]
\centering
\begin{subfigure}[b]{0.27\textwidth}
    \centering
    \includegraphics[width=\linewidth]{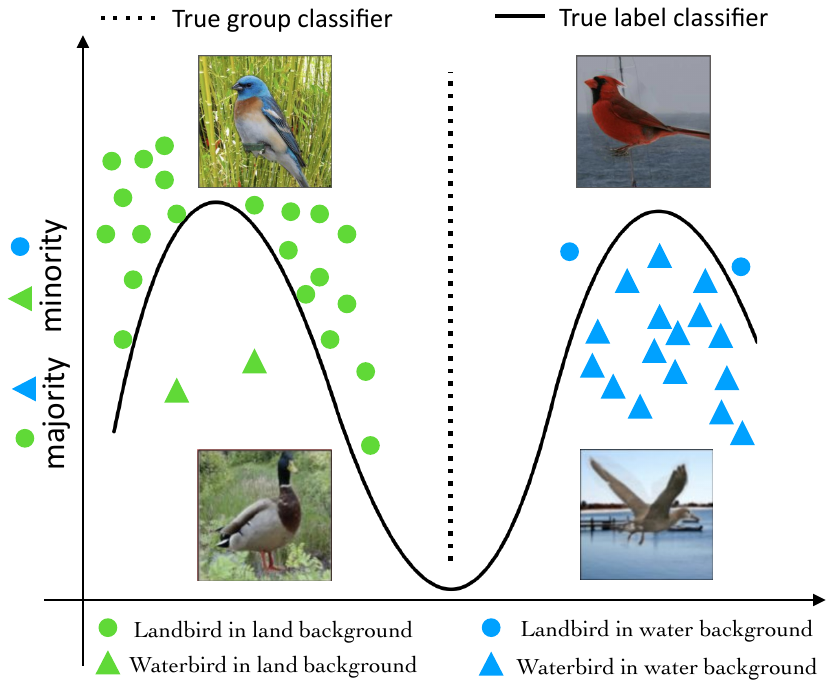}
  \end{subfigure} 
  \begin{subfigure}[b]{0.55\textwidth}
    \centering
    \includegraphics[width=\linewidth]{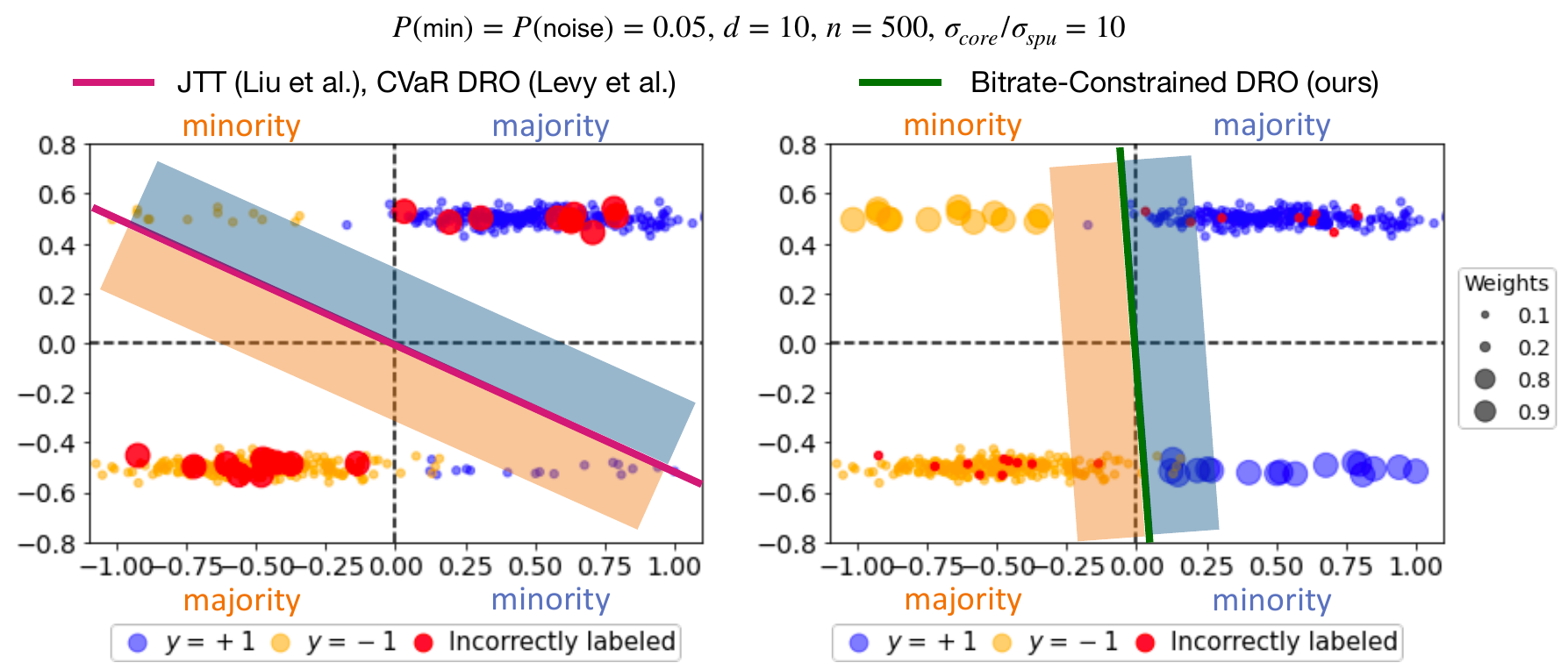}
  \end{subfigure}
\caption{\footnotesize 
\textbf{Bitrate-Constrained DRO}: A method that assumes group shifts along low-bitrate features, and restricts the adversary appropriately so that the solution found is less pessimistic and more robust to unknown group shifts. Our method is also robust to training noise.
\emph{(Left)} In Waterbirds~\citep{wah2011caltech}, 
the spurious feature background is a large margin simple feature that
separates the \emph{majority} and \emph{minority} points in each class. \emph{(Right)} Prior works~\citep{levy2020large,liu2021just} that upweight arbitrary points with high losses force the model to memorize noisy mislabeled points while our method is robust to noise and only upweights the true minority group without any knowledge of its identity (see Section~\ref{subsec:rcn}).
 }
 \label{fig:intro-figure}
\vspace{-1.5em}
\end{figure}

Prior works consider various approaches to address group shift. One solution is to ensure robustness to worst case shifts using distributionally robust optimization (DRO)~\citep{bagnell2005robust,ben2013robust,duchi2016statistics}, which considers a two-player game where a {learner} minimizes risk on distributions chosen by an {adversary} from a predefined uncertainty set. 
As the adversary is only constrained to propose distributions that lie within an f-divergence based uncertainty set,
DRO often yields overly conservative (pessimistic) solutions~\citep{hu2018does} and can suffer from statistical challenges~\citep{duchi2019distributionally}. This is mainly because DRO upweights high loss points that may not form a meaningful group in the real world, and may even be \emph{contrived} if the high loss points simply correspond to randomly mislabeled examples in the training set.   
Methods like \gdro ~\citep{sagawa2019distributionally} avoid overly pessimistic solutions by assuming knowledge of group membership for each training example. However, these group-based methods provide no guarantees on shifts that deviate from the predefined groups (\eg when there is a new group), 
and are not applicable to problems that lack group knowledge. In this work, we therefore ask: \textit{Can we train non-pessimistic robust models without access to group information on training samples?}
We address this question by considering a more nuanced assumption on the structure of the underlying groups. 
We assume that, conditioned on the label, group boundaries are realized by high-level features 
that depend on a small set of underlying factors (\eg background color, brightness). 
This leads to simpler group functions with large margin and simple decision boundaries between groups (Figure~\ref{fig:intro-figure} \emph{(left)}).
Invoking the principle of minimum description length~\citep{grunwald2007minimum}, restricting our adversary to functions that satisfy this assumption corresponds to a bitrate constraint. 
In DRO, the adversary upweights points with higher losses under the current learner, which in practice often correspond to examples that belong to a rare group, contain complex patterns, or are mislabeled~\citep{carlini2019distribution,toneva2018empirical}. 
Restricting the adversary's capacity prevents it from upweighting individual hard or mislabeled examples (as they cannot be identified with simple features), and biases it towards identifying erroneous data points misclassified by simple features.
This also complements the failure mode of neural networks trained with stochastic gradient descent (SGD) that rely on simple spurious features which correctly classify points in the \emph{majority} group but may fail on \emph{minority} groups ~\citep{blodgett2016demographic}.

The main contribution of this paper is Bitrate-Constrained DRO (\bdro), 
a supervised learning procedure that provides robustness to distribution shifts  along groups realized by simple functions.
Despite not using group information on training examples, we demonstrate that \bdro can match the performance of methods requiring them. 
We also find that \bdro is more successful in identifying true minority training points, compared to unconstrained DRO.
This indicates that not optimizing for performance on contrived worst-case shifts can reduce the pessimism inherent in DRO. 
It further validates: (i) our assumption on the simple nature of group shift; 
and (ii) that our bitrate constraint meaningfully structures the uncertainty set to be robust to such shifts. 
As a consequence of the constraint, we also find that \bdro is robust to random noise in the training data~\citep{song2022learning}, since it cannot form ``groups'' entirely based on randomly mislabeled points with low bitrate features. This is in contrast with existing methods that use the learner's training error to up-weight arbitrary sets of difficult training points~\citep[\eg][]{liu2021just,levy2020large}, which we show are highly susceptible to label noise (see Figure~\ref{fig:intro-figure}~\emph{(right)}).
Finally, we theoretically analyze our approach---characterizing how the degree of constraint on the adversary can effect worst risk estimation and excess risk (pessimism) bounds, as well as convergence rates for specific online solvers.

%% file: sections/relwork.tex
\section{Related Work}
\label{sec:relwork}


Prior works in robust ML~\citep[e.g.,][]{li2018learning,lipton2018detecting,goodfellow2014explaining} address various forms of adversarial or structured shifts. We specifically review prior work on robustness to group shifts. While those based on DRO  
optimize for worst-case shifts in an explicit uncertainty set, the robust set is implicit for some others, with most using some form of importance weighting.

\textbf{Distributionally robust optimization (DRO).} 
DRO methods generally optimize for worst-case performance on joint $(\rvx, \ry)$ distributions that lie in an $f$-divergence ball (uncertainty set) around the training distribution~\citep{ben2013robust,rahimian2019distributionally,bertsimas2018data,blanchet2019quantifying,miyato2018virtual,duchi2016statistics,duchi2021uniform}.
\citet{hu2018does} highlights that the conservative nature of DRO may lead to degenerate solutions when the unrestricted adversary uniformly upweights all misclassified points. \citet{sagawa2019distributionally} proposes to address this by limiting the adversary to shifts that only differ in marginals over predefined groups. However, in addition to it being difficult to obtain this information, 
\citet{kearns2018preventing} raise ``gerrymandering'' concerns with  notions of robustness that fix a small number of groups apriori. While they propose a solution that looks at exponentially many subgroups defined over protected attributes, our method does not assume access to such attributes and aims to be fair on them as long as they are realized by simple functions.
Finally, \citet{zhai2021boosted} avoid conservative solutions by solving the DRO objective over randomized predictors learned through boosting. We consider deterministic and over-parameterized learners and instead constrain the adversary's class.   

\textbf{Constraining the DRO uncertainty set.} 
In the marginal DRO setting, \citet{duchi2019distributionally} limit the adversary via easier-to-control reproducing kernel hilbert spaces (RKHS) or bounded H\"{o}lder continuous functions \citep{liu2014robust,wen2014robust}.
While this reduces the statistical error in worst risk estimation, the size of the uncertainty set (scales with the data)  remains too large to  avoid cases where an adversary can re-weight mislabeled and hard examples from the majority set~\citep{carlini2019distribution}.
In contrast, we  restrict the adversary even for large datasets where the estimation error would be low,
as this would reduce excess risk when we only care about robustness to rare sub-populations defined by simple functions. Additionally, while their analysis and method prefers the adversary's objective to have a strong dual, we show empirical results on real-world datasets and generalization bounds where the adversary's objective is not necessarily convex. 

\textbf{Robustness to group shifts without demographics.} 
Recent works~\citep{sohoni2020no,creager2021environment,bao2022learning} that aim to achieve group robustness without access to group labels employ various heuristics where the robust set is implicit while others require data from multiple domains~\citep{arjovsky2019invariant,yao2022improving} or ability to query test samples~\citep{lee2022diversify}. 
\citet{liu2021just} use training losses for a heavily regularized model trained with empirical risk minimization (ERM) to directly identify minority data points with higher losses and re-train on the dataset that up-weights the identified set. \citet{nam2020learning} take a similar approach. Other methods~\citep{idrissi2022simple} propose simple baselines that subsample the majority class in the absence of group demographics and the majority group in its presence. 
\citet{hashimoto2018fairness} find DRO over a $\chi^2$-divergence ball can reduce the otherwise increasing disparity of per-group risks in a dynamical system. Since it does not use features to upweight points (like \bdro) it is vulnerable to label noise. Same can be said about some other works (\eg \cite{liu2021just,nam2020learning}).   

\textbf{Importance weighting in deep learning.} Finally, numerous works~\citep{duchi2016statistics,levy2020large,lipton2018detecting,oren2019distributionally} enforce robustness by re-weighting losses on individual data points. Recent investigations~\citep{soudry2018implicit,byrd2019effect,lu2022importance} reveal that such objectives have little impact on the learned solution in interpolation regimes. One way to avoid this pitfall is to train with heavily regularized models~\citep{sagawa2019distributionally, sagawa2020investigation} and employ early stopping. Another way is to subsample certain points, as opposed to up-weighting~\citep{idrissi2022simple}. In this work, we use both techniques while training our objective and the baselines, ensuring that the regularized class is robust to shifts under misspecification~\citep{wen2014robust}. 

%% file: sections/preliminaries.tex
\section{Preliminaries}
\label{sec:prelim}

We introduce the notation we use in the rest of the paper and describe the DRO problem. In the following section, we will formalize our assumptions on the nature of the shift before introducing our optimization objective and algorithm.  

\textbf{Notation.} With covariates $\gX \subset \Real^d$ and labels $\gY$, the given source $P$ and unknown true target $Q_0$ are measures over the measurable space $(\gX \times \gY, \Sigma)$ and have densities $p$ and $q_0$ respectively (w.r.t. base measure $\mu$). 
The learner's choice is a hypothesis $h: \gX \mapsto \gY$ in class $\gH \subset L^2(P)$, and the adversary's action in standard DRO is a target distribution $Q$ in set $\gQ_{P, \kappa} \coloneqq  \{Q :  Q \ll P,\,  D_{f}(Q\, ||\, P) \leq \kappa\}$. 
Here, $D_f$ is the $f$-divergence between $Q$ and $P$ for a convex function $f$\footnote{For \eg $\kl{Q}{P}$ can be derived with $f(x) = x \log x$ and for Total Variation $f(x) = |x-1|/2$.} with $f(1)=0$. An equivalent action space for the adversary is the set of re-weighting functions:
 \begin{align}
    \gW_{P,\kappa} = \{w: \gX \times \gY \mapsto \Real \st w \; \textrm{is measurable under}\; P, \; \E_{P}[w] = 1,\; \E_P{f(w)} \leq \kappa \}
    \label{eq:adv-constraints}
\end{align}
For a convex loss function $l: \gY \times \gY \mapsto \Real_+$, we denote $l(h)$ as the function over $(\rvx, \ry)$ that evaluates $l(h(\rvx), \ry)$, and use $\lzone$ to denote the loss function $\I(h(\rvx) \neq \ry)$.
Given either distribution $Q \in \gQ_{P, \kappa}$, 
or a re-weighting function $w \in \gW_{P, \kappa}$, the risk of a learner $h$ is:
\begin{align}
& \;\; \;\;\;R(h, Q)  =  \E_{Q}\;[l(h)]  \;\;\;\;\;\;\;\;\; R(h, w) =  \E_{(\rvx, \ry) \sim P} \; [l(h(\rvx), \ry) \cdot w(\rvx, \ry)] =  \innerprod{l(h)}{\;w}_P
\label{eq:risk-defn} 
\end{align}
Note the overload of notation for $R(h, \cdot)$. If the adversary is stochastic it picks a mixed action $\delta \in \dwp$, which is the set of all distributions over $\gW_{P, \kappa}$. Whenever it is clear, we drop $P, \kappa$.

\textbf{Unconstrained DRO~\citep{ben2013robust}.} This is a min-max optimization problem understood as a two-player game, where the learner chooses a hypothesis, to minimize risk on the worst distribution that the adversary can choose from its set.
Formally, this is given by \eqref{eq:dro-main}.
The first equivalence is clear from the definitions and for the second since $R(h, Q)$ is linear in $Q$, the supremum over $\dwp$ is a Dirac delta over the best weighting in $\gW_{P, \kappa}$. In the next section, we will see how a bitrate-constrained adversary can only pick certain actions from $\dwp$.
\begin{align} 
 & \inf_{h \in \gH} \; \sup_{Q \in \gQ_{P, \kappa}} \; R(h, Q)\;\; \equiv \;\; \inf_{h \in \gH} \; \sup_{w \in \gW_{P, \kappa}} \; R(h, w) \;\; \equiv \;\; \inf_{h \in \gH} \; \sup_{\delta \in \dwp} \;\; \E_{w \sim \delta} \brck{R(h, w)} 
\label{eq:dro-main}
\end{align}
\textbf{Group Shift.} While the DRO framework in Section~\ref{sec:prelim} is broad and addresses any unstructured shift, we focus on the specific case of group shift. 
First, for a given pair of measures $P, Q$ we define what we mean by the group structure $\gG_{P, Q}$ (Definition~\ref{def:group-struc}). Intuitively, it is a set of sub-populations along which the distribution shifts, defined in a way that makes them uniquely identifiable. For \eg in the Waterbirds dataset (Figure~\ref{fig:intro-figure}), there are four groups given by combinations of (label, background). Corollary~\ref{cor:group-struc-unique} follows immediately from the definition of $\gG_{P, Q}$. Using this definition, the standard group shift assumption~\citep{sagawa2019distributionally} can be formally re-stated as Assumption~\ref{assm:group-shift}. 
\begin{definition}[group structure $\gG_{P, Q}$]
    \label{def:group-struc}
    For $Q$ $\ll$ $P$ the group structure $\gG_{P, Q}$$=$$\{G_k\}_{k=1}^{K}$ is the smallest finite set of disjoint groups $\{G_k\}_{k=1}^K$ s.t. $Q(\cup_{k=1}^K G_k)$$=$$1$ and $\forall$$k$ (i) $G_k \in \Sigma$, $Q(G_k) > 0$ and (ii) $p(\rvx,\ry \mid G_k) = q(\rvx,\ry \mid G_k) > 0 \; a.e$. in $\mu$. If such a structure exists then $\gG_{P, Q}$ is well defined.  
\end{definition}
\begin{corollary}[uniqueness of $\gG_{P, Q}$]
    $\forall P, Q$, the structure $\gG(P, Q)$ is unique if it is well defined.
    \label{cor:group-struc-unique}
\end{corollary}
\begin{assumption}[standard group shift]
    \label{assm:group-shift}
    There exists a well-defined group structure $\gG_{P, Q_0}$ s.t. target $Q_0$ differs from $P$ only in terms of marginal probabilities over all $ G \in \gG_{P, Q_0}$.
\end{assumption}


%% file: sections/bDRO.tex
\section{Bitrate-Constrained DRO}
\label{sec:bdro}

We begin with a note on the expressivity of the adversary in Unconstrained DRO and  formally introduce the assumption we make on the nature of shift. Then, we build intuition for why unconstrained adversaries fail but restricted ones do better under our assumption. 
Finally, we state our main objective and discuss a specific instance of it. 

\textbf{How expressive is unconstrained adversary?}
Note that the set $\gW_{P, \kappa}$ includes all measurable functions (under $P$) such that the re-weighted distribution is bounded in $f$-divergence (by $\kappa$). 
While prior works~\citep{shafieezadeh2015distributionally,duchi2016statistics}  shrink $\kappa$ 
to construct confidence intervals,
this \textit{only controls} the total mass that can be moved between measurable sets $G_1, G_2 \in \Sigma$, but \textit{does not restrict} the choice of $G_1$ and $G_2$ itself. As noted by \citet{hu2018does}, such an adversary is highly expressive, and optimizing for the worst case only leads to the solution of empirical risk minimization (ERM) under $\lzone$ loss. Thus, we can conclude that DRO recovers degenerate solutions because the worst target in $\gW_{P,\kappa}$ lies far from the subspace of naturally occurring targets. Since it is hard to precisely characterize natural targets we make a nuanced assumption: the target $Q_0$ only upsamples those rare subpopulations that are misclassified by simple features. We state this  formally in Assumption~\ref{assm:simple-group-shift} after we define the bitrate-constrained function class $\gW(\gamma)$ in Definition~\ref{def:bitrate-constrained-class}. 
\begin{definition}
    \label{def:bitrate-constrained-class} A function class $\gW(\gamma)$ is bitrate-constrained if there exists a data independent prior $\pi$, s.t. $\gW(\gamma) =  \{\E[\delta] \st \delta \in \Delta(\gW), \; \kl{\delta}{\pi} \leq \gamma\}$.
\end{definition}
 
\begin{assumption}[simple group shift]
    \label{assm:simple-group-shift}
    Target $Q_0$ satisfies Assumption~\ref{assm:group-shift} (group shift) w.r.t. source $P$. 
    Additionally,     
    For some prior $\pi$ and a small $\gamma^*$, the re-weighting function $q_0/p$ lies in a bitrate-constrained class $\gW(\gamma^*)$. In other words, for every group $G \in \gG(P, Q_0)$, $\exists w_G \in \gW(\gamma^*)$ s.t. $\I((\rvx, \ry) \in G) = w_G$ a.e.. We refer to such a $G$ as a \textbf{simple group} that is realized in $\gW(\gamma^*)$.
\end{assumption}
Under the principle of minimum description length ~\citep{grunwald2007minimum} any deviation from the prior (\ie $\kl{\delta}{\pi}$) increases the \emph{description length} of the encoding $\delta \in \Delta(\gW)$, thus we refer to $\gW(\gamma)$ as being \emph{bitrate-constrained} in the sense that it contains functions (means of distributions) that can be described with a limited number of bits given the prior $\pi$. See Appendix~\ref{subsec:assm-explain} for an example of a bitrate-constrained class of functions. Next we present arguments for why identifiability of simple (satisfy Assumption~\ref{assm:simple-group-shift}) minority groups can be critical for robustness.  
\textbf{Neural networks can perform poorly on simple minorities.}
For a fixed target $Q_0$, let's say there exists two groups: $\gmin$ and $\gmaj \in \gG(P, Q_0)$ such that $P(\gmin) \ll P(\gmaj)$. 
By Assumption~\ref{assm:simple-group-shift}, both $\gmin$ and $\gmaj$ are simple (realized in $\gW(\gamma^*)$), and are thus separated by some simple feature. 
The learner's class $\gH$ is usually a class of overparameterized neural networks. When trained with stochastic gradient descent (SGD), these are biased towards learning simple features that classify a majority of the data~\mbox{\citep{shah2020pitfalls,soudry2018implicit}}. Thus, if the simple feature separating $\gmin$ and $\gmaj$ itself correlates with the label $y$ on $\gmaj$, then neural networks would fit on this feature. This is precisely the case in the Waterbirds example, where the groups are defined by whether the simple feature background correlates with the label (Figure~\ref{fig:intro-figure}). Thus our assumption on the nature of shift complements the nature of neural networks perform poorly on simple minorities.



\textbf{The bitrate constraint helps identify simple unfair minorities in $\gG(P, Q_0)$.} Any method that aims to be robust on $Q_0$ must up-weight data points from $\gmin$ but without knowing its identity.
Since the unconstrained adversary upsamples any group of data points with high loss and low probability, it cannot distinguish between a rare group that is realized by simple functions in $\gW(\gamma^*)$ and a rare group of examples that share no feature in common or may even be mislabeled. On the other hand, the group of mislabeled examples cannot be separated from the rest by functions in $\gW(\gamma^*)$. Thus, a bitrate constraint adversary can only identify simple groups and upsamples those that incur high losses -- possibly due to the simplicity bias of neural networks. 

\textbf{\bdro objective.} According to Assumption~\ref{assm:simple-group-shift}, there cannot exist a target $Q_0$ such that minority $\gmin \in \gG(P, Q_0)$ is not realized in bitrate constrained class $\gW(\gamma^*)$.
Thus, by constraining our adversary to a class $\gW(\gamma)$ (for some $\gamma$ that is user defined), we can possibly evade issues emerging from optimizing for performance on mislabeled or hard examples, even if they were rare. This gives us the objective in Equation~\ref{eq:bdro-main} where the equalities hold from the linearity of $\innerprod{\cdot}{\cdot}$ and Definition~\ref{def:bitrate-constrained-class}.
\begin{align}
    \inf_{h \in \gH}  \sup_{\substack{\delta \in \Delta(\gW)  \\ \kl{\delta}{\pi} \leq \gamma}}  \E_{w \sim \delta} R(h, w)   \; = \;   
    \inf_{h \in \gH}  \sup_{\substack{\delta \in \Delta(\gW)  \\ \kl{\delta}{\pi} \leq \gamma}}   \langle l(h), \E_\delta[w] \rangle_P   \;\; = \;\;   \inf_{h \in \gH}  \sup_{w \in \wgam}  R(h, w) \label{eq:bdro-main} 
\end{align}
\textbf{\bdro in practice.} We parameterize the learner $\thh \in \Theta_h$ and adversary $\thw \in \Theta_w$ as neural networks\footnote{We use $\theta_h, \theta_w$ and $l(\theta_h)$ to denote $w(\thw; (\rvx, \ry)), h(\thh; \rvx)$ and $l(h(\thh; \rvx), \ry)$ respectively.}.
In practice, we implement the adversary either as a one hidden layer variational information bottleneck (VIB)~\citep{alemi2016deep}, where the Kullback-Leibler (KL) constraint on the latent variable $\rvz$ (output of VIB's hidden layer) directly constrains the bitrate; or as an $l_2$ norm constrained linear layer. 
The objective for the VIB ($l_2$) version is obtained by setting  $\bvib \neq 0$ ($\bltwo \neq 0$) in \eqref{eq:bdro-prac} below. See Appendix~\ref{appsubsec:bdro-objective} for details.
Note that the objective in Equation~\ref{eq:bdro-prac} is no longer convex-concave  and can have multiple local equilibria or stationary points~\citep{mangoubi2021greedy}. The adversary's objective also does not have a strong dual that can be solved through conic programs---a standard practice in DRO literature~\citep{namkoong2016stochastic}. Thus, we provide an algorithm where both learner and adversary optimize \bdro iteratively through stochastic gradient ascent/descent (Algorithm~\ref{alg:online-bdro} in Appendix~\ref{subsec:bdro-algo}). 
\begin{align}
        & \;\;\; \fourquad \min_{\thh \in \Theta_h} \innerprod{l({\thh})}{\thws}_P \;\;\;\; \textrm{s.t.} \;\;\;\; \thws = \argmax_{\thw \in \Theta_w} \;\;L_{\textrm{adv}}(\thw; \btheta_h, \bvib, \bltwo, \eta) \label{eq:bdro-prac}   \\
        & L_{\textrm{adv}}(\thw; \btheta_h, \bvib, \bltwo, \eta) =  \innerprod{l({\thh}) - \eta}{\thw}_P - \bvib \; \E_{P} \kl{p(\rvz \;|\; \rvx; \thw)}{\gN(\bf{0}, {I}_d)} - \beta_{l_2} \|\thw\|_2^2 \nonumber
\end{align}
\textbf{Training.} For each example, the adversary takes as input: (i) the last layer output of the current learner's feature network; and (ii) the input label. The adversary then outputs a weight (in $[0, 1]$). The idea of applying the adversary directly on the learner's features (instead of the original input) is based on recent literature \citep{rosenfeld2022domain,kirichenko2022last} that suggests re-training the prediction head is sufficient for robustness to shifts. The adversary tries to maximize weights on examples with value $\geq \eta$ (hyperparameter) and minimize on others. 
For the learner, in addition to the example it takes as input the adversary assigned weight for that example from the previous round and uses it to reweigh its loss in a minibatch. Both players are updated in a round (Algorithm~\ref{alg:online-bdro}). 

%% file: sections/theory.tex
\section{Theoretical Analysis}
\label{sec:analysis}
\newcommand{\etaD}{\hat{\eta}_D^\gamma}

The main objective of our analysis of \bdro is to show how adding a bitrate constraint on the adversary can: (i) give us tighter statistical estimates of the worst risk; and (ii) control the pessimism (excess risk) of the learned solution. First, we provide worst risk generalization guarantees using the PAC-Bayes framework~\citep{catoni2007pac}, along with a result for kernel adversary. Then, we provide convergence rates and pessimism guarantees for the solution found by our online solver for a specific instance of $\gW(\gamma)$
For both these, we analyze the constrained form of the conditional value at risk (CVaR) DRO objective~\citep{levy2020large} below.

\textbf{Bitrate-Constrained CVaR DRO.} When the uncertainty set $\gQ$ is defined by the set of all distributions $Q$ that have bounded likelihood \ie $\|q/p\|_\infty \leq 1/\alpha_0$, we recover the original CVaR DRO objective\mbox{~\citep{duchi2021uniform}}. The bitrate-constrained version of CVaR DRO is given in \eqref{eq:bitcon-cvar-dro} (see Appendix~\ref{appsec:omitted-proofs} for derivation). Note that, slightly different from Section~\ref{sec:prelim}, we define $\gW$ as the set of all measurable functions $w$$:$ $\gX$$\times$ $\gY$ $\mapsto$ $[0,1]$, since the other convex restrictions in \eqref{eq:adv-constraints} are handled by dual variable $\eta$. As in Section~\ref{sec:bdro}, $\gW(\gamma)$ is derived from $\gW$ using Definition~\ref{def:bitrate-constrained-class}.  In \eqref{eq:bitcon-cvar-dro}, if we replace the bitrate-constrained class $\gW(\gamma)$ with the unrestricted $\gW$ then we recover the variational form of unconstrained CVaR DRO in \citet{duchi2016statistics}. 
\begin{align}
\label{eq:bitcon-cvar-dro} 
     \gL^*_{\textrm{cvar}}(\gamma)  =  \inf_{h \in \gH, \eta \in \Real}  \sup_{w \in \gW(\gamma)}  R(h, \eta, w)\;\;\;\;\;\;\;\;\textrm{where,}\;\;\;\;\;\;
    R(h, \eta, w) = (1/\alpha_0) \innerprod{l(h)-\eta}{w}_P + \eta
\end{align} 
\textbf{Worst risk estimation bounds for \bdro.} Since we are only given a finite sampled dataset $\gD$ $\sim$ $P^n$, we solve the objective in \eqref{eq:bitcon-cvar-dro} using the empirical distribution $\pn$. We denote the plug-in estimates as $\hD^\gamma, \etaD$. 
This incurs an estimation error for the true worst risk. 
But when we restrict our adversary to $\dwgam$, for a fixed learner $h$ we reduce the worst-case risk estimation error which scales with the bitrate $\kl{\cdot}{\pi}$ of the solution (deviation from prior $\pi$). Expanding this argument to every learner in $\gH$,  
with high probability we also reduce the estimation error for the worst risk of $\hD^\gamma$. Theorem~\ref{thm:worst-risk-gen} states this  generalization guarantee more precisely.
\begin{theorem}[worst-case risk generalization]
With probability $\geq 1-\delta$ over $\gD \sim P^n$, the worst bitrate-constrained $\alpha_0$-CVaR risk for $\hD^\gamma$ can be upper bounded by the following oracle inequality:
{
\footnotesize
\begin{align}
     \sup_{w \in \wgam} R(\hD^\gamma, \etaD, w) \;\lsim \; \gL^*_{\textrm{cvar}}(\gamma) +  \frac{M}{\alpha_0}  \sqrt{\paren{\gamma + \log\paren{\frac{1}{\delta}} + (d+1) \log\paren{\frac{L^2n}{\gamma}} +  \log n}/{(2n -1)}} \nonumber,
\end{align}}
when $l(\cdot, \cdot)$ is $[0,M]$-bounded, $L$-Lipschitz and $\gH$ is parameterized by convex set $\Theta \subset \Real^d$.
    \label{thm:worst-risk-gen}
\end{theorem}

Informally, Theorem~\ref{thm:worst-risk-gen} tells us that bitrate-constraint $\gamma$ gracefully controls the estimation error $\gO(\sqrt{(\gamma + \gC(\gH))/n})$ (where $\gC(\gH)$ is a complexity measure) if we know that Assumption~\ref{assm:simple-group-shift} is satisfied. While this only tells us that our estimator is consistent with $\gO_p(1/\sqrt{n})$, the estimate may itself be converging to a degenerate predictor, \ie $ \gL^*_{\textrm{cvar}}(\gamma)$ may be very high. For example, if the adversary can cleanly separate mislabeled points even after the bitrate constraint, then presumably these noisy points with high losses would be the ones mainly contributing to the worst risk, and up-weighting these points would result in a learner that has memorized noise. Thus, it becomes equally important for us to analyze the excess risk (or the pessimism) for the learned solution. Since this is hard to study for any arbitrary bitrate-constrained class $\wgam$, we shall do so for the specific class of reproducing kernel Hilbert space (RKHS) functions.

\textbf{Special case of bounded RKHS.} 
Let us assume there exists a prior $\Pi$ such that $\wgam$ in Definition~\ref{def:bitrate-constrained-class} is given by an RKHS induced by  Mercer kernel $k:\gX\times\gX \mapsto\Real$, s.t. the eigenvalues of the kernel operator decay polynomially, \ie $\mu_j \lsim j^{{-2}/{\gamma}}$ $(\gamma < 2)$. Then, if we solve for $\hD^\gamma,\etaD$ by doing kernel ridge regression over norm bounded  ($\|f\|_{\wgam}$$\leq$$B\leq1$) smooth functions $f$ then we can control: (i) the pessimism of the learned solution; and (ii) the generalization error (Theorem~\ref{thm:special-case-rkhs}). Formally, we refer to pessimism for estimates $\hD^\gamma, \etaD$ as excess risk defined as:
{
\footnotesize
\begin{align}
    \label{eq:excess-risk}
    \textrm{excess risk} \coloneqq \sup_{w \in \wgam}|\inf_{h,\eta} R(h, \eta, w) - R(\hD^\gamma, \etaD, w)|. 
\end{align}}
\begin{theorem}[bounded RKHS]
    \label{thm:special-case-rkhs} For $l, \gH$ in Theorem~\ref{thm:worst-risk-gen}, and for $\wgam$ described above $\exists$$\gamma_0$ s.t. for all sufficiently bitrate-constrained $\gW(\gamma)$ \ie $\gamma$$\leq$$\gamma_0$, w.h.p.  $1- \delta$ worst risk generalization error is 
    $\gO\paren{(1/n)\paren{\log(1/\delta) + (d+1) \log(nB^{-\gamma} L^{\gamma/2})}}$ 
    and the excess risk is $\gO(B)$ for $\hD^\gamma, \etaD$ above. 
\end{theorem}
Thus, in the setting described above we have shown how bitrate-constraints given indirectly by $\gamma, R$ can control both the pessimism and statistical estimation errors. Here, we directly analyzed the estimates $\hD^\gamma, \etaD$ but did not describe the specific algorithm used to solve the objective in \eqref{eq:bitcon-cvar-dro} with $\pn$. Now, we look at an iterative online algorithm to solve the same objective and see how bitrate-constraints can also influence convergence rates in this setting. 

\textbf{Convergence and excess risk analysis for an online solver.} In the following, we provide an algorithm to solve the objective in \eqref{eq:bitcon-cvar-dro} and analyze how bitrate-constraint impacts the solver and the solution.
For convex losses, the min-max objective in \eqref{eq:bitcon-cvar-dro} has a unique solution and this matches the unique Nash equilibrium for the generic online algorithm (game) we describe (Lemma~\ref{lem:nash-eq}). The algorithm is as follows: Consider a two-player zero-sum game where the learner uses a no-regret strategy to first play $h \in \gH, \eta \in \Real$ to minimize $\E_{w \sim \delta} R(h, \eta, w)$. Then, the adversary plays follow the regularized leader (FTRL) strategy to pick distribution $\delta \in \Delta(\wgam)$ to maximize the same. 
Our goal is to analyze the bitrate-constraint $\gamma$'s effect on the above algorithm's convergence rate and the pessimistic nature of the solution found. For this, we need to first characterize the bitrate-constraint class $\gW(\gamma)$. 
If we assume there exists a prior $\Pi$ such that $\wgam$ is Vapnik-Chervenokis (VC) class of dimension $O(\gamma)$, then in Theorem~\ref{thm:convergence-excess-guarantee}, we see that the iterates of our algorithm converge to the equilibrium (solution) in $\gO({\sqrt{{\gamma\log n}/{T}}})$ steps. Clearly, the degree of bitrate constraint can significantly impact the convergence rate for a generic solver that solves the constrained DRO objective. 
Theorem~\ref{thm:convergence-excess-guarantee} also bounds the excess risk (\eqref{eq:excess-risk}) on $\pn$.
\begin{lemma}[Nash equilibrium]
\label{lem:nash-eq} For strictly convex $l(h)$, $l(h) \in [0,M]$, the objective in \eqref{eq:bitcon-cvar-dro} has a unique solution which is also the Nash equilibrium of the game above when played over compact sets $\gH \times [0, M]$, $\dwgam$. We denote this equilibrium as $h^*_D(\gamma), \eta^*_D(\gamma), \delta^*_D(\gamma)$.
\end{lemma}
\begin{theorem}
\label{thm:convergence-excess-guarantee}    
At time step $t$, if the learner plays $(h_t, \eta_t)$ with no-regret and the adversary plays $\delta_t$ with FTRL strategy that uses a negative entropy regularizer on $\delta$ 
then average iterates $(\bar{h}_{T},\bar{\eta}_{T},\bar{\delta}_{T}) =(1/T) \sum_{t=1}^T (h_t, \eta_t, \delta_t)$ converge to the equilibrium $(h^*_D(\gamma), \eta^*_D(\gamma), \delta^*_D(\gamma))$ at rate $\gO({\sqrt{{\gamma\log n}/{T}})}$. Further the excess risk defined above is $ \gO((M/\alpha_0)\paren{1-\frac{1}{n^\gamma}})$. 
\end{theorem}

%% file: sections/experiments.tex
\vspace{-0.7em}
\section{Experiments}
\label{sec:experiments}
\vspace{-0.7em}

Our experiments aim to evaluate the performance of \bdro and compare it with ERM and group shift robustness methods that do not require group annotations for training examples. We conduct empirical analyses along the following axes: (i) worst group performance on datasets that exhibit known spurious correlations; (ii) robustness to random label noise
in the training data; (iii) average performance on hybrid covariate shift datasets with unspecified groups; and (iv) accuracy in identifying minority groups. See Appendix~\ref{appsec:additional-expts} for additional experiments and details\footnote{The code used in our experiments can be found at \url{https://github.com/ars22/bitrate_DRO}.}.


\textbf{Baselines.} Since our objective is to be robust to group shifts without  group annotations on training examples,
we explore baselines that either optimize for the worst minority group (CVaR DRO~\citep{levy2020large}) or use training losses to identify specific minority points (LfF~\citep{nam2020learning}, JTT~\citep{liu2021just}).   
\gdro~\citep{sagawa2019distributionally} is treated as an oracle. 
We also compare with the simple re-weighting baseline (RWY) proposed by \citet{idrissi2022simple}.

\textbf{Implementation details.} We train using Resnet-50~\citep{he2016deep} for all methods and datasets except CivilComments, where we use BERT~\citep{wolf2019huggingface}. For our VIB adversary, we use a $1$-hidden layer neural network encoder and decoder (one for each label). As mentioned in Section~\ref{sec:bdro}, the adversary takes as input the learner model's features and the true label to generate weights. All implementation and design choices for baselines were adopted directly from \citet{liu2021just,idrissi2022simple}. We provide model selection methodology and other details in Appendix~\ref{appsec:additional-expts}.

\newcommand{\mt}[1]{{\footnotesize{(#1)}}}
\newcommand{\mbl}[1]{#1}
\newcommand{\mbb}[1]{#1}
\begin{table}[!t]
    \vspace{-2em}
    \scriptsize
    \centering
    \setlength{\tabcolsep}{1em}
    \begin{tabular}{r|cccccc}
          & \multicolumn{2}{c}{Waterbirds} & \multicolumn{2}{c}{CelebA} & \multicolumn{2}{c}{CivilComments}   \\ 
            Method    & Avg  & WG & Avg  & WG & Avg & WG \\ \midrule
    ERM                               & 97.1 \mt{0.1} &       71.0 \mt{0.4} & 95.4 \mt{0.2} &       46.9 \mt{1.0} & 92.3 \mt{0.2} & 57.2 \mt{0.9} \\
    LfF ~\citep{nam2020learning}      & 90.7 \mt{0.2} &       77.6 \mt{0.5} & 85.3 \mt{0.2} &       77.4 \mt{0.7} & 92.4 \mt{0.1} & 58.9 \mt{1.1} \\
    RWY ~\citep{idrissi2022simple}    & 93.7 \mt{0.3} &       85.8 \mt{0.5} & 84.9 \mt{0.2} &       80.4 \mt{0.3} & 91.7 \mt{0.2} & 67.7 \mt{0.7} \\
    JTT ~\citep{liu2021just}          & 93.2 \mt{0.2} &       86.6 \mt{0.4} & 87.6 \mt{0.2} &       81.3 \mt{0.5} & 90.8 \mt{0.3} & 69.4 \mt{0.8} \\ 
    CVaR DRO ~\citep{levy2020large}   & 96.3 \mt{0.2} & \mbl{75.5 \mt{0.4}} & 82.2 \mt{0.3} & \mbl{64.7 \mt{0.6}} & 92.3 \mt{0.2} & \mbl{60.2 \mt{0.8}} \\\midrule
    \bdro (\vib) (ours)               & 94.1 \mt{0.2} & \mbb{86.3 \mt{0.3}} & 86.7 \mt{0.2} & \mbb{80.9 \mt{0.4}} & 90.5 \mt{0.2} & \mbb{68.7 \mt{0.9}} \\ 
    \bdro (\ltwop) (ours)             & 93.8 \mt{0.2} & \mbb{86.4 \mt{0.3}} & 87.7 \mt{0.3} & \mbb{80.4 \mt{0.6}} & 91.0 \mt{0.3} & \mbb{68.9 \mt{0.7}} \\ \midrule \gdro
  ~\cite{sagawa2019distributionally}  & 93.2 \mt{0.3} & 91.1 \mt{0.3} & 92.3 \mt{0.3} & 88.4 \mt{0.6} & 88.5 \mt{0.3} & 70.0 \mt{0.5}               
    \end{tabular}
    \vspace{-0.5em}
    \caption{\footnotesize \textbf{\bdro recovers worst group performance gap between CVaR DRO and Group DRO:} On Waterbirds, CelebA and CivilComments  we report test average (Avg) and test worst group (WG) accuracies for \bdro and baselines. In ($\cdot$) we report the standard error of the mean accuracy across five runs.}
    \label{tab:sc-datasets}
    \vspace{-1em}
\end{table}

\textbf{Datasets.}  For experiments in the known groups and label noise settings we use:  (i) Waterbirds~\citep{wah2011caltech} (background is spurious), CelebA~\citep{liu2015deep} (binary gender is spuriously correlated with label ``blond''); and  CivilComments (WILDS)~\citep{borkan2019nuanced} where the task is to predict ``toxic'' texts and there are 16 predefined groups~\cite{koh2021wilds}. We use FMoW and Camelyon17~\citep{koh2021wilds} to test methods on datasets that do not have explicit group shifts. In FMoW the task is to predict land use from satellite images where the training/test set comprises of data before/after 2013. Test involves both subpopulation shifts over regions (\eg Africa, Asia) and domain generalization over time (year). Camelyon17 presents a domain generalization problem where the task is to detect tumor in tissue slides from different sets of hospitals in train and test sets. 

\vspace{-0.5em}
\subsection{Is \bdro robust to group shifts without training data group annotations?}
\label{subsec:known-correlations}
\vspace{-0.5em}

Table~\ref{tab:sc-datasets} compares the average and worst group accuracy for \bdro with ERM and four group shift robustness baselines: JTT, LtF, SUBY, and CVaR DRO. First, we see that unconstrained CVaR DRO underperforms other heuristic algorithms. This matches the observation made by \citet{liu2021just}. Next, we see that adding bitrate constraints on the adversary via a KL term or \ltwop penalty significantly improves the performance of \bdro (\vib) or \bdro (\ltwop),
which now matches the best performing baseline (JTT). Thus, we see the less conservative nature of \bdro allows it to recover a large portion  of the performance gap between \gdro and CVaR DRO. Indirectly, this partially validates our Assumption~\ref{assm:simple-group-shift}, which states that the minority group is identified by a low bitrate adversary class. In Section~\ref{subsec:identify-min} we  discuss exactly what fraction of the minority group is identified, and the role played by the strength of bitrate-constraint.

\begin{figure}[t]
    \centering
    \vspace{-2.2em}
    \begin{subfigure}[b]{0.2\textwidth}
        \centering
        \includegraphics[width=\linewidth]{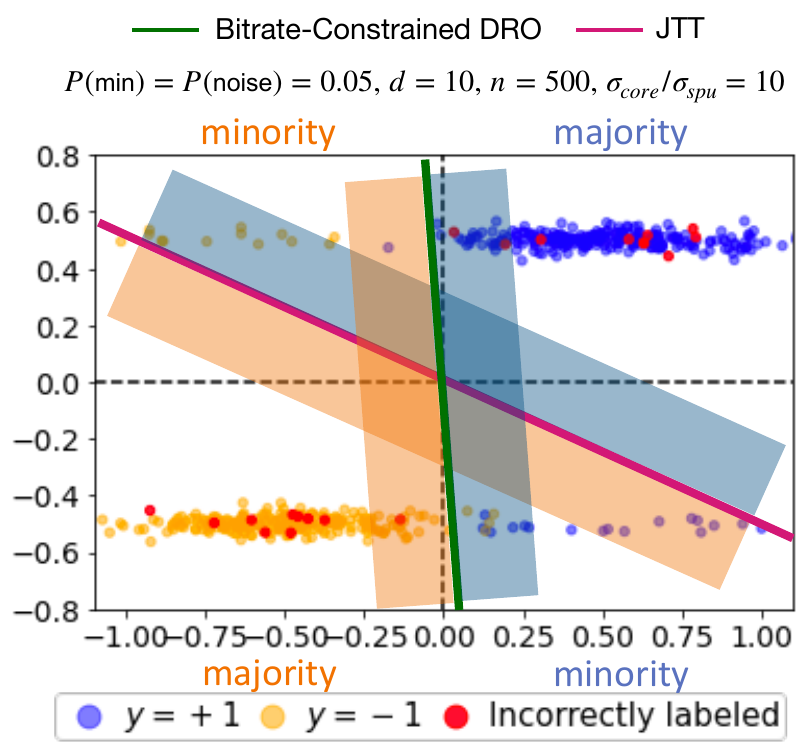}
        \caption{\label{fig:noise-robustness-left}}
    \end{subfigure}
    \hspace{2em}
    \begin{subfigure}[b]{0.68\textwidth}
        \centering
        \includegraphics[width=\linewidth]{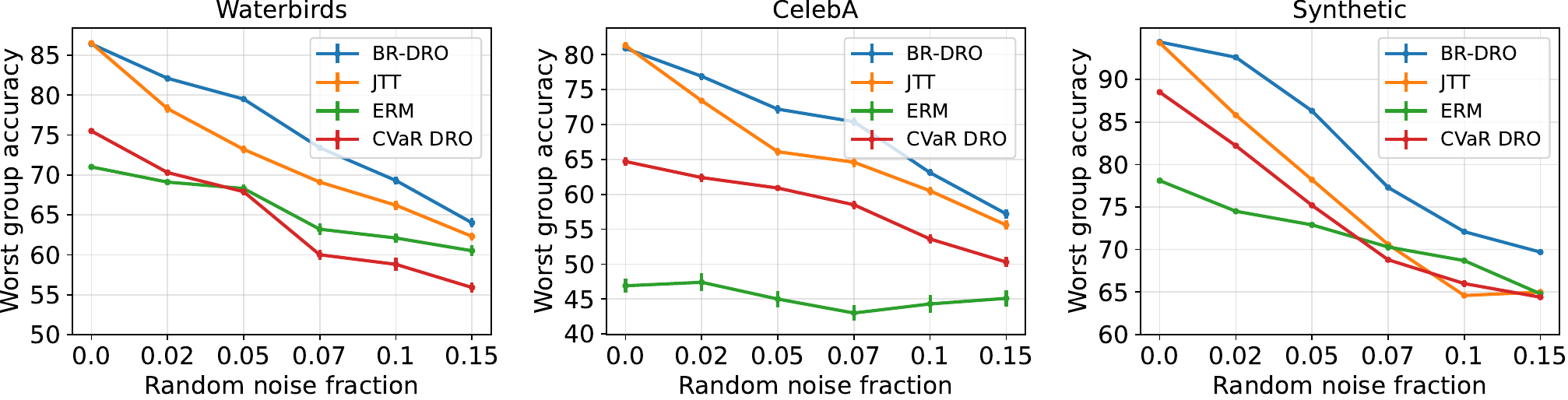}    
        \caption{\label{fig:noise-robustness-right}}
    \end{subfigure}
        \vspace{-0.5em}
    \caption{\footnotesize
    \emph{(Left)} \textbf{Visualization (2d) of noisy synthetic data and learned predictors:} We plot the decision boundaries (projected onto core and spurious features) learned by JTT and \bdro when the adversary is restricted to a sparse predictor. While our method recovers the core feature the baselines memorize the minority points.
    \emph{(Right)} \textbf{\bdro is robust to random label noise in training data:} Across varying levels of noise fraction in training data we compare performance of \bdro with ERM and methods (JTT, CVaR DRO) that naively up weight high loss points.\vspace{-2em}}
\end{figure}

\vspace{-0.5em}
\subsection{\bdro is more robust to random label noise}
\vspace{-0.5em}
\label{subsec:rcn}
Several methods  for group robustness (\eg CVaR DRO, JTT) are based on the idea of up weighting points with high training losses. The goal is to obtain a learner with matching
performance on every (small) fraction of points in the dataset. However, when training data has mislabeled examples, such an approach will likely yield degenerate solutions. This is because the adversary directly upweights any example where the learner has high loss, including datapoints with incorrect labels. Hence, even if the learner's prediction matches the (unknown) true label, this formulation would force the learner to {memorize} incorrect labelings at the expense of learning the true underlying function.  
On the other hand, if the adversary is sufficiently bitrate constrained,  it cannot upweight the arbitrary set of randomly mislabeled points, as this would require it to memorize those points. Our Assumption~\ref{assm:simple-group-shift} also dictates that the distribution shift
would not upsample such high bitrate noisy examples. 
Thus, our constraint on the adversary ensures \bdro is robust to label noise in the training data and our assumption on the target distribution retains its robustness to test time distribution shifts.  


In Figure~\ref{fig:noise-robustness-right} we highlight this failure mode of unconstrained up-weighting methods
in contrast to \bdro. We first induce random label noise~\citep{carlini2019distribution} of varying degrees into the Waterbirds and CelebA training sets.
Then we run each method and compare worst group performance. In the absence of noise we see that the performance of JTT is comparable with \bdro, if not slightly better (Table~\ref{tab:sc-datasets}). Thus, both \bdro and JTT perform reasonably well in identifying and upsampling the simple minority group in the absence of noise. In its presence, \bdro significantly outperforms JTT and other approaches on both Waterbirds and CelebA, as it only upsamples the minority examples misclassified by simple features, ignoring the noisy examples for the reasons above. 
To further verify our claims, we set up a noisily labeled synthetic dataset (see Appendix~\ref{appsec:additional-expts} for details).
In Figure~\ref{fig:noise-robustness-left} we plot training samples 
as well as the solutions learned by \bdro and and JTT on synthetic data. In Figure~\ref{fig:intro-figure}\emph{(right)} we also plot exactly which points are upweighted by \bdro and JTT. Using both figures, we note that JTT mainly upweights the noisy points (in red) and memorizes them using $\xnoise$. Without any weights on minority, it memorizes them as well and learns component along spurious feature.
On the contrary, when we restrict the adversary with \bdro to be sparse ($l_1$ penalty),
it only upweights minority samples, since no sparse predictor can separate noisy points in the data. Thus, the learner can no longer memorize the upweighted minority and we recover the robust predictor along core feature.  





\vspace{-0.25em}
\subsection{What fraction of minority is recovered by \bdro?}
\label{subsec:identify-min}
\begin{figure}[t!]
    \centering
    \vspace{-2em}
    \includegraphics[width=0.95\linewidth]{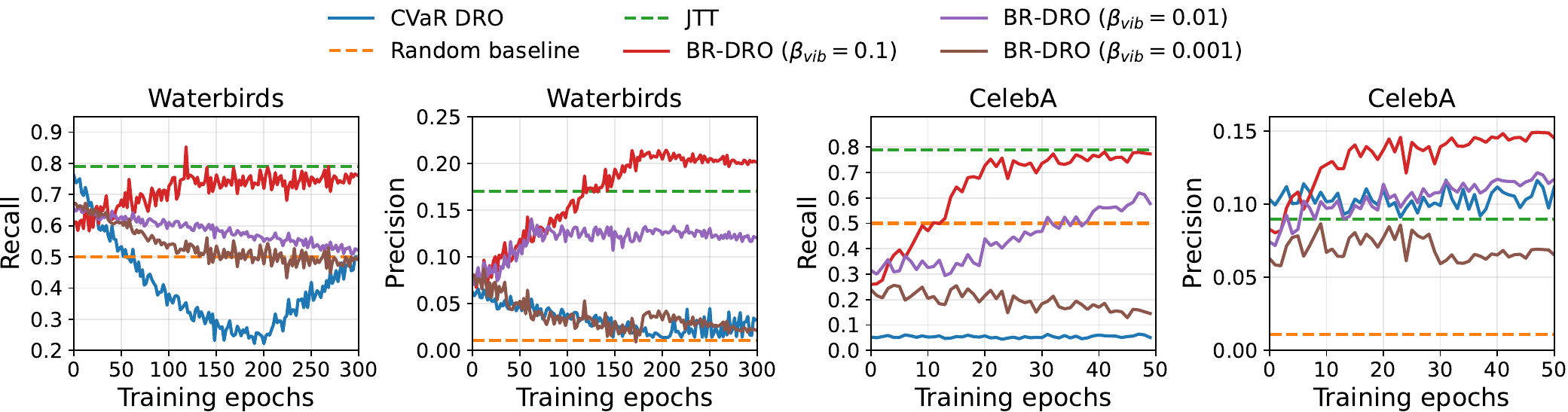}
    \vspace{-0.5em}
    \caption{\footnotesize By considering the fraction of points upweighted by our adversary (top $10\%$) as the positive class we analyze the precision and recall of this class with respect to the minority group.  and do the same for JTT, random baseline and CVaR DRO. \bdro achieves highest precision and matches recall with \jtt asymptotically. We also find that increasing bitrate constraint $\bvib$ helps improving precision/recall.}
    \label{fig:precision-recall}
    \vspace{-1.5em}
\end{figure}

\vspace{-0.1em}
We claim that our less pessimistic objective can more accurately recover (upsample) the true minority group if indeed the minority group is simple (see Assumption~\ref{assm:simple-group-shift} for our definition of simple).
In this section, we aim to verify this claim. If we treat examples in the top $10\%$ (chosen for post hoc analysis) fraction of examples as our predicted minorities, we can check precision and recall of this decision on the Waterbirds and CelebA datasets. Figure~\ref{fig:precision-recall} plots these metrics at each training epoch for \bdro (with varying $\bvib$), \jtt and CVaR DRO. Precision of the random baseline tells us the true fraction of minority examples in the data. First we note that \bdro consistently performs much better on this metric than unconstrained CVaR DRO. In fact, as we reduce strength of $\bvib$ we recover precision/recall close to the latter. This controlled experiment shows that the bitrate constraint is helpful (and very much needed) in practice to identify rare simple groups. 
In Figure~\ref{fig:precision-recall} we observe that asymptotically, the precision of \bdro is better than \jtt on both datasets, while the recall is similar. Since importance weighting has little impact in later stages with exponential tail losses~\citep{soudry2018implicit, byrd2019effect}, other losses (\eg polytail~\citet{wang2021importance}) may further improve the performance of \bdro as it gets better at identifying the minority classes when trained longer.

\vspace{-0.5em}
\subsection{How does \bdro perform on more general covariate shifts?}
\label{subsec:gen-cov-shift}
\vspace{-0.5em}

In Table~\ref{tab:gen-cov-shifts} we report the average test accuracies for \bdro and baselines on the hybrid dataset FMoW and domain generalization dataset Camelyon17. 
Given its hybrid nature, on FMoW we also report worst region accuracy. First, we note that on these datasets group shift robustness baselines do not do better than ERM. Some are either too pessimistic (\eg CVaR DRO), or require heavy assumptions 
\begin{wraptable}{r}{0.53\textwidth}
        \scriptsize
        \centering
        \setlength\tabcolsep{0.2pt}
        \setlength{\tabcolsep}{0.45em}
    \begin{tabular}{r|cc|c}
         Method & \multicolumn{2}{c}{FMoW} & \multicolumn{1}{c}{Camelyon17}   \\ 
                & Avg  & W-Reg & Avg      \\ \midrule
            ERM                              & 53.3 \mt{0.1} & 32.4 \mt{0.3} & 70.6 \mt{1.6} \\ \midrule
            JTT ~\cite{liu2021just}          & 52.1 \mt{0.1} & 31.8 \mt{0.2} & 66.3 \mt{1.3} \\
            LfF ~\cite{nam2020learning}      & 49.6 \mt{0.2} & 31.0 \mt{0.3} & 65.8 \mt{1.2} \\
            RWY ~\cite{idrissi2022simple}    & 50.8 \mt{0.1} & 30.9 \mt{0.2} & 69.9 \mt{1.3} \\
Group DRO ~\cite{sagawa2019distributionally} & 51.9 \mt{0.2} & 30.4 \mt{0.3} & 68.5 \mt{0.9} \\          
            CVaR DRO ~\cite{levy2020large}   & 51.5 \mt{0.1} & 31.0 \mt{0.3} & 66.8 \mt{1.3} \\ \midrule
            \bdro (\vib) (ours)              & 52.0 \mt{0.2} & 31.8 \mt{0.2} & 70.4 \mt{1.5} \\ 
            \bdro (\ltwop) (ours)            & 53.1 \mt{0.1} & 32.3 \mt{0.2} & 71.2 \mt{1.0} \\ 
    \end{tabular}
    \vspace{-0.5em}
    \caption{\label{fig:wilds-dataset} \footnotesize Average (Avg) and worst region (W-Reg for FMoW) test accuracies on Camelyon17 and FMoW.\vspace{-0.1em}
    \label{tab:gen-cov-shifts}
    }
\end{wraptable}
(\eg Group DRO) to be robust to domain generalization.
 This is also noted by~\citet{gulrajani2020search}. Next, we see that \bdro (\ltwop version) does better than other group shift baselines
on both both worst region and average  datasets and matches ERM performance on Camelyon17. One explanation could be that even though these datasets test models on new domains, there maybe some latent groups defining these domains that are simple and form a part of latent subpopulation shift. Investigating this claim further is a promising line of future work. 

%% file: sections/conclusion.tex
\vspace{-0.35em}
\section{Conclusion}
\label{sec:conclusion}
\vspace{-1em}
In this paper, we proposed a method for making machine learning models more robust. While prior methods optimize robustness on a per-example or per-group basis, our work focuses on features. In doing so, we avoid requiring group annotations on training samples, but also avoid the excessively conservative solutions that might arise from CVaR DRO with fully unconstrained adversaries. Our results show that our method avoids learning spurious features, is robust to noise in the training labels, and does better on other forms of covariate shifts compared to prior approaches. Our theoretical analysis also highlights other provable benefits in some settings like reduced estimation error, lower excess risk and faster convergence rates for certain solvers.
\vspace{-0.2em}

\textbf{Limitations.}
While our method lifts the main limitation of Group DRO (access to training group annotations), it does so at the cost of increased complexity. Further, to tune hyperparameters, like prior work we assume access to a some group annotations on validation set but also get decent performance (on some datasets) with only a balanced validation set (see Appendix~\ref{appsec:additional-expts}). Adapting group shift methods to more generic settings remains an important and open problem.

{
\textbf{Acknowledgement.} The authors would like to thank Tian Li, Saurabh Garg at Carnegie Mellon University, and Yoonho Lee at Stanford University for helpful feedback and discussion. 
}



%% file: sections/appendix.tex
\clearpage

\section*{Appendix Outline}


\ref{appsec:practical-imp} Implementing \bdro in practice

\ref{appsec:additional-expts} Additional empirical results and other experiment details

\ref{appsec:omitted-proofs} Omitted Proofs.



\input{sections/practical}

\input{sections/additional_exp}
\input{sections/proofs}

%% file: sections/practical.tex
\section{Implementing \bdro in practice}
\label{appsec:practical-imp}

\subsection{\bdro algorithm}
\label{subsec:bdro-algo}
If the bitrate constraint is applied via the KL term in \eqref{eq:bdro-prac}, we implement the adversary as a variational information bottleneck~\citep{alemi2016deep} (\vib), where the KL divergence with respect to a standard Gaussian prior controls the bitrate of the adversary's feature set $\rvz \sim p(\rvz\mid \rvx; \thw)$. 
Increasing $\bvib$ can be seen as enforcing lower bitrate features \ie reducing $\gamma$ in $\wgam$
(smaller value of $\kl{\delta}{\pi}$ in the primal formulation in Definition~\ref{def:bitrate-constrained-class}). 
If the constraint is applied via the $l_2$ term we implement the adversary as a linear layer. In some cases (\eg Section~\ref{subsec:rcn}) we use a sparsity constraint ($l_1$ norm) on the linear adversary.

\SetKwComment{acp}{/* }{ */}
\begin{algorithm}[!h]
 \DontPrintSemicolon
 \KwIn{Adversary VIB penalty $\bvib$; Step sizes $\eta_l, \eta_w$; Dataset $\gD=(\rvx_i, \ry_i)_{i=1}^{n}$} 
 Initialize $\theta_h^{(1)}$ and $\theta_w^{(1)}$\;
 \For{$t=1,\dots,T$}{
  From $\gD$, sample $ \rvx, \ry \sim \gD$ \acp*{Sample datapoint} 
  $\btheta_h^{(t+1)} \gets \Pi_{\Theta_h} \paren{\btheta_h^{(t)} - \eta_h  \nabla_{\thh} \brck{ l(\btheta_h^{(t)}(\rvx), \ry) \cdot \thw(\rvx, \ry)}}$ \acp*{Update $\btheta_h$}
  $\btheta_w^{(t+1)} \gets \Pi_{\Theta_w} \paren{\btheta_w^{(t)} + \eta_w  \nabla_{\thw} L_{\textrm{adv}}(\btheta_w^{(t)}; \btheta_h^{(t)}, \bvib, \bltwo, \eta) }$ \acp*{Update $\btheta_w$}
 }
 \KwOut{${\bar{\btheta}_h} = \frac{1}{T} \sum_{t=1}^T \btheta_h^{(t)}$, ${\bar{\btheta}_w} = \frac{1}{T} \sum_{t=1}^T \btheta_w^{(t)}$}
 \caption{Bitrate-Constraint DRO (Online Algorithm)}
 \label{alg:online-bdro}
\end{algorithm}

\subsection{\bdro objective in Equation~\ref{eq:bdro-prac}}
\label{appsubsec:bdro-objective}
When describing the actual \bdro objective in~\eqref{eq:bdro-prac}, for brevity we used $\thh$ to denote both the parameters of the learner and the learner itself (similarly for $\thw$). Here, we describe in detail the parameterized version of the objective in~\eqref{eq:bdro-main}, and clarify the modeling of the adversary.

Let us denote the learner as $h(\cdot; \thh):\gX \mapsto \gY$, and the class of learners $\gH \coloneqq \{h(\cdot; \thh) \st \thh \in \Theta_h\}$. A learner is composed of two parts: i) a feature extractor $f:\gX \mapsto \Real^{p}$ that maps the input into a $p$-dimensional feature vector and; ii) a classifier $c:\Real^p \mapsto \gY$ that maps the features into a predicted label. 
Similarly, in the bitrate-constrained class $\wgam$ each adversary is parameterized as $w(\cdot; \thw): \Real^p \times \gY \mapsto [0, 1]$ where with some overloading of notation we use $w(\rvx, \ry)$ to denote $w(f(\rvx), \ry)$ \ie the adversary only operates on the features output by the feature extractor $f(\rvx)$ and the given label $\ry$\footnote{
Note that, constraining the output space of the adversary to be bounded $([0, 1])$ does not necessarily deviate from our definition of $\gW_{P,\kappa}$ in \eqref{eq:adv-constraints}, since we can always re-define the output space by dividing the $w \in \gW_{P,\kappa}$ by $\max_{\vx, y} w(\vx , y),\, \forall w  \in \gW_{P,\kappa}$.}. Since the features are extracted by the deep neural network (and frozen), the adversary is implemented as either a two layer neural network with VIB constraint or a linear layer with $l_2$ constraint -- both of which promote low bitrate functions satisfying our Assumption~\ref{assm:simple-group-shift}. We shall now describe each version of \bdro separately.

\textbf{\bdro (VIB)}: For each label $\ry$, $w(f(\rvx), \ry)$ is a one hidden layer neural network with ReLU activations (1-layer VIB). The first layer takes as input feature vector $f(\rvx)$ and outputs a $2d$ dimensional vector $\rvu$. The $d$ dimensional latent encoding $\rvz_\thw(\rvx)$ (dependence on $\rvx$ is made explicit) is sampled from multivariate Normal $\rvz_\thw(\rvx) \sim \gN(\rvu_\thw[\textrm{:d}], \mathrm{diag}(\rvu_\thw[\textrm{d:}]))$ where $\rvu_\thw[\textrm{:d}]$ is the mean and $\mathrm{diag}(\rvu_\thw[\textrm{d:}])$ is a diagonal covariance matrix, both parameterized by parameter $\thw$ (neural net). Following~\cite{alemi2016deep}, an information bottleneck constraint is applied on the latent variable $\rvz_\thw(\rvx)$ in the form of a  KL constraint with respect to standard Gaussian prior, with strength given by scalar $\bvib$. Prior works~\citep{tishby2015deep,hjelm2018learning} have argued why this regularization would bias the adversary to learn low bitrate functions. If we assume the generative model for $\rvx$ as one defined by latent factors of variation (\eg orientation, background), then presumably the group identity is a function of these factors. Finally, an output layer with a sigmoid activation maps $\rvz_\thw(\rvx)$ into a weight between $[0, 1]$.
If we believe that the KL constraint on $\rvz_\thw(\rvx)$ helps recover some of these factors, then learning a linear transform (with sigmoid activation) over it would amount to learning a simple group function.

\textbf{\bdro ($l_2$)}: For each label $\ry$, $w(f(\rvx), \ry)$ is a linear layer. It takes as input feature vector $f(\rvx)$ and maps it to a scalar which when passed through a sigmoid yields a weight between $[0, 1]$. The $l_2$ constraint over the linear layer parameters is controlled by scalar $\bltwo$. This corresponds to bitrate constraints under certain priors \citep{polson2019bayesian}.
We can incorporate the above parameterizations for VIB and $l_2$ versions of \bdro into the \bdro objective in \eqref{eq:bdro-main} through \eqref{eq:bdro-prac-vib} and \eqref{eq:bdro-prac-l2} respectively to yield the final objective in~\eqref{eq:bdro-prac}. Note that as we mention in Section~\ref{sec:bdro}, we can switch between the two versions of \bdro by setting $\bvib=0$ (for $l_2$) or $\bltwo=0$ (for VIB). While we can choose to constrain the adversary with both forms of constraints simultaneously we find that in practice picking only one of them for a given problem instance helps with tuning the degree of constraint. Finally, while $\bvib, \bltwo$ act as Lagrangian parameters for our bitrate constraint, $\eta$ is the Lagrangian parameter for the constrains on $w(\rvx, \ry)$ in the definition of $\wgam_{P, \kappa}$ in \eqref{eq:adv-constraints}. 
\begin{align}
        & \;\;\; \fourquad \min_{h(\cdot; \thh) \in \gH} \;\;\;\; \E_{P} \brck{l(h(\rvx;\; \thh), \ry) \cdot w^*(f(\rvx), \ry;\; \thws)} \nonumber \\
        & \textrm{s.t.} \;\;\;\; w^* = \argmax_{w(\cdot; \thw) \in \wgam} \;\;L_{\textrm{adv}}(\thw; \btheta_h, \bvib, \bltwo, \eta)   \nonumber \\
        & L_{\textrm{adv}}(\thw, \thh, \bvib, \bltwo, \eta) = \E_{P} \brck{\paren{l(h(\rvx;\; \thh), \ry)- \eta} \cdot w(f(\rvx), \ry;\; \thw)} \nonumber \\
        & \fourquad - \bvib  \cdot  \E_{\rvx \sim P(\rvx)} \brck{ \kl{\rvz_\thw(\rvx)}{\gN(\bf{0}, {I}_d)} }  \label{eq:bdro-prac-vib} \\
        & \fourquad - \beta_{l_2} \cdot \|\thw\|_2^2  \label{eq:bdro-prac-l2}
\end{align}


\subsection{Connecting bitrate-constrained \texorpdfstring{$\wgam$}{wgamma} to simple groups (a practical example).}
\label{subsec:assm-explain}

First, let us recall the definition of a simple group in Assumption~\ref{assm:simple-group-shift}. We defined a group $G$ to be simple, if the indicator function $\I((\rvx, \ry) \in G)$ identifying said group is containing in the bitrate-constrained class of functions $W(\gamma)$. Next, we shall see what this means in terms of a specific prior $\pi$ and the constraint $\kl{\cdot}{\pi} \leq \gamma$ that defines the class $W(\gamma)$ in Definition~\ref{def:bitrate-constrained-class}.

Let us assume that the adversary $(\thw)$ is parameterized as a linear classifier in $\Real^d$ where each parameter $\thw$ corresponds to a re-weighting function in $\gW_{P,\kappa}$ (\eqref{eq:adv-constraints}), and the prior $\pi$ is designed to have a higher likelihood over low norm solutions. Specifically, $\pi$ takes the form: 
$$\pi(\thw) \propto \exp{(-\|\thw\|_2^2)}.$$ 
Now consider the class of densities: 
$$ \Delta(\Theta_w) \coloneqq \{\delta_{\thw}(\btheta)  \propto \exp{(-\|\btheta - \thw \|_2^2)} \st \thw \in \Theta_w \}.$$ 
Here, it is easy to verify that: 
$$\E_{\btheta \sim \delta_{\thw}} [\btheta] = \thw.$$  
Note, that while Definition~\ref{def:bitrate-constrained-class} concerns with any $\delta \in \Delta(\gW)$, we restrict ourselves to the subset $\Delta(\Theta_w) \subset \Delta(\gW)$ for the sake of mathematical convenience, and find that even this set is rich enough to easily violate the bitrate-constraint as we shall see next. 

Applying Definition~\ref{def:bitrate-constrained-class} on the set $\Delta(\Theta_w)$ we get: 
$$\wgam = \{\thw \st \kl{\delta_\thw}{\pi} \leq \gamma\}.$$
Further if we compute $\kl{\delta_\thw}{\pi}$, we find that $\kl{\delta_\thw}{\pi} = \norm{\thw}_2^2 + C$ (for some constant C). Thus, the bitrate constraint  $\kl{\delta_\thw}{\pi} \leq \gamma$ directly transfers into an $l_2$ norm constraint on the mean parameter $\thw$, and $\wgam$ is simply the set of parameters $\thw \in \Theta_w$ that have their $l_2$ norms bounded above by some constant $\gamma - C$. The objective for the $l_2$ version of our adversary in \eqref{eq:bdro-prac} reflects this form. Hence, this example connects norm constrained parametric adversaries to bitrate constrained simple group identity functions.

%% file: sections/additional_exp.tex
\section{Additional empirical results and other experiment details}
\label{appsec:additional-expts}

\subsection{Hyper-parameter tuning methodology}

There are two ways in which we tune hyperparameters on datasets with known groups (CelebA, Waterbirds, CivilComments): (i) on average validation performance; (ii) worst group accuracy. The former does not use group annotations while the latter does. Similar to prior works~\cite{liu2021just,idrissi2022simple} we note that using group annotations (on a small validation set) does improve performance. In Table~\ref{tab:hyp} we report our study which varies the the fraction $p$ of group labels that are available at test time. For each setting of $p$, we do model selection by taking weighted (by $p$) mean over two entities (i) average validation on all samples, (ii) worst group validation on a fraction $p$ of minority samples. In the case where $p=0$, we only use average validation. We report our results on CelebA and Waterbirds dataset. For the two WILDS datasets we tune hyper-parameters on OOD Validation set.

\begin{table}[!ht]
    \centering
    \footnotesize
    \setlength{\tabcolsep}{0.25em}
    \begin{tabular}{r|cccc|cccc}
         & \multicolumn{4}{c}{Waterbirds} & \multicolumn{4}{c}{CelebA}  \\ 
                 Method & $p=0.0$ & $p=0.02$ & $p=0.05$ & $p=0.1$ & $p=0.0$ & $p=0.02$ & $p=0.05$ & $p=0.1$  \\\midrule
JTT                     & 62.7   & 73.9    & 77.3   & 84.4          & 42.1   & 68.3    & 80.5   & 80.3  \\
CVaR DRO                & 63.9   & 65.8    & 72.6   & 74.1          & 33.6   & 40.4    & 60.4   & 63.2  \\ 
LfF                     & 48.6   & 58.9    & 70.3   & 79.5          & 34.0   & 58.9    & 60.0   & 78.3  \\ 
\bdro (\vib)            & \textbf{69.3}   & 77.6    & 76.1   & 84.9 & \textbf{52.4}   & 71.2    & 80.3   & 79.9  \\
\bdro (\ltwop)          & \textbf{68.9}   & 75.2    & 79.4   & 86.1 & \textbf{55.8}   & 63.5    & 74.6   & 80.4 \\
    \end{tabular}
    \caption{We check to what extent fraction of group annotations in the training data affect performance. For each dataset and method, we tune its hyper-parameters on the average validation and worst group (only on the small fraction $p$ that is available). We see that while all methods consistently improve as we increase group annotations and tune for worst group accuracy on the annotated samples, \bdro does do better that prior works when tuned on just average validation ($p=0.$). At the same time, we note that this still does not match the performance of \bdro when tuned on worst group validation (seen in Table~\ref{tab:sc-datasets}).}
    \label{tab:hyp}
\end{table}

\subsection{Synthetic dataset details}

We follow the explicit-memorization setup in \citet{sagawa2020investigation} which we summarize here briefly. Let input $\rvx = [\xcore, \xspu, \xnoise]$ where $\xcore \mid y \sim \gN(y, \sigcore^2), \; \xspu \mid a \sim \gN(a, \sigspu^2)$ and $\xnoise \sim \gN(\bf{0}, (\signoise^2 \bf{I}_d)/d )$. Here $a \in \{-1, 1\}$ refers to a spurious attribute, and label is $y \in \{-1, 1\}$,
We set $a = y$ with probability $P(\textrm{\tiny maj})=1-P(\textrm{\tiny min})$. The level of correlation between $a$ and $y$ is controlled by $P(\textrm{\tiny maj})$. Additionally, we flip true label with probability $P(\textrm{\tiny noise})$.

\subsection{Degree of constraint}

\begin{figure}[!ht]
    \centering
    \includegraphics[width=0.6\linewidth]{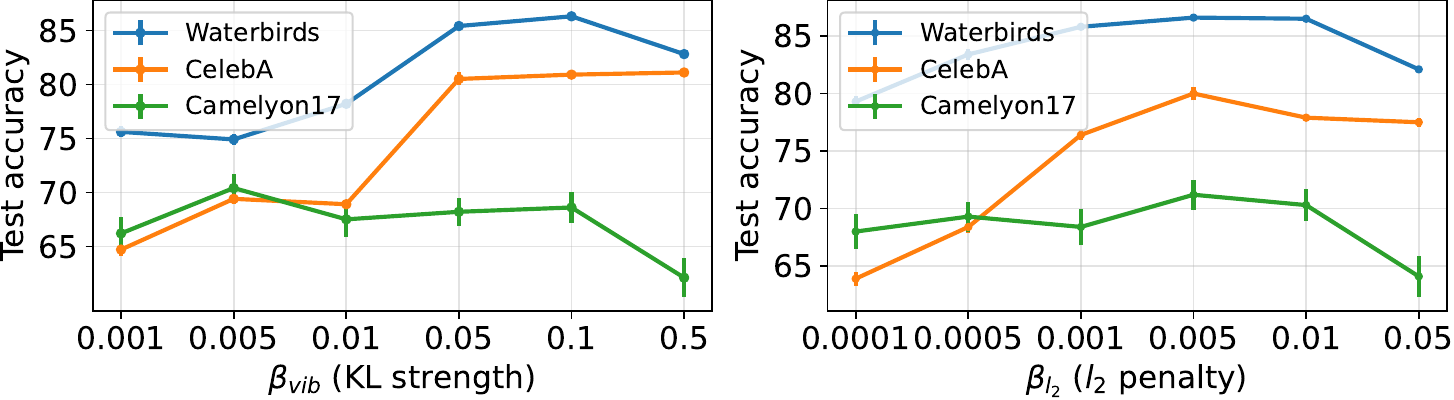}
    \caption{\textbf{Optimal bitrate-constraints for robustness to distributions shifts:} For two versions of capacity control: KL, $l_2$ penalty (see Section~\ref{sec:bdro}) we show how worst group performance on Waterbirds, CelebA and average performance on Camelyon test sets improves with increasing constraints under either VIB ($\bvib$) or linear ($\bltwo$) adversaries.\vspace{-0.5em}}
    \label{fig:beta-ablation}
\end{figure}

In Figure~\ref{fig:beta-ablation} we see how worst group performance varies on Waterbirds and CelebA as a function of increasing constraint. We also plot average performance on the Camelyon dataset. We mainly note that for either of the constraint implementations, only when we significantly increase the capacity do we actually see the performance of \bdro improve. The effect is more prominent on groups shift datasets with simple groups (Waterbirds, CelebA).  Under less restrictive capacity constraints we note that its performance is similar to CVaR DRO (see Figure~\ref{fig:precision-recall}). This is expected since CVaR DRO is the completely unconstrained version of our objective. 

\subsection{Hyper-parameter details.} 

For all hyper-parameters of prior methods we use the ones state in their respective prior works. The implementation \gdro, \jtt, CVaR DRO is borrowed from the implementation made public by authors of \citet{liu2021just}. For datasets Waterbirds, CelebA and CivilComments we choose the hyper-parameters (whenever applicable) learning rate, batch size, weight decay on learner, optimizer, early stopping criterion, learning rate schedules used by \citet{liu2021just} for their implementation of CVaR DRO method. For datasets FMoW and Camelyon17 we choose values for these hyper-parameters to be the ones used by \citet{koh2021wilds} for the ERM baseline. Details on \bdro specific hyper-parameters that we tuned are in Table~\ref{tab:hyp-details}. Also, note that we release our implementation with this submission. 

\begin{table}[!ht]
    \footnotesize
    \centering
    \begin{tabular}{c|c|c|c|c|c}
        Hyper-parameter & Waterbirds & CelebA & CivilComments & FMoW & Camelyon17 \\ \midrule
        learning rate for adversary & 0.01 & 0.05 & 0.001 & 0.02 & 0.01 \\
        threshold  $\eta$  & 0.05 & 0.05 & 0.1 & 0.1 & 0.1  \\ 
        $\bvib$ & 0.1 & 0.1 & 0.02 & 0.005 & 0.005 \\ 
        $\bltwo$ & 0.01 & 0.005 & 0.005 & 0.02 & 0.005 \\ 
    \end{tabular}
    \caption{Hyperparameters for our method on different datasets (tuned on worst group validation performance). Note, that the threshold $\eta$ here is the top $x\%$ fraction.}
    \label{tab:hyp-details}
\end{table}

\subsection{Fine-grained evaluation of worst-case performance on CivilComments.}
\label{appsubsec:civil-comments-fine-grained}

The CivilComments dataset~\citet{koh2021wilds} is a collection of comments from online articles, where each comment is rated for its toxicity, In addition, there is information available on $8$ demographic identities: \textit{male, female, LGBTQ, Christian, Muslim, other religions, Black and White} for each comment, \ie a given comment may be attributed to one or more of these demographic identities~\citet{koh2021wilds}. There are 133,782 instances in the test set. Each of the $8$ identities form two groups based on the toxicity label, for a total of 16 groups. These are the groups used in the training of methods that assume group knowledge like Group DRO and also used in the evaluation. In Table~\ref{tab:sc-datasets} we report the accuracy over the worst group (on the test set) for each of the methods.  
Here, we do a more fine-grained evaluation of the worst-group performance. In addition to the $16$ groups that methods are typically evaluated on, we evaluate the performance over groups created by combinations of two different demographic identities and a label (e.g., (\textit{male}, \textit{christian}, \textit{toxic}) or (\textit{female}, \textit{Black}, \textit{not toxic})). Thus, the spurious attribute is no longer binary since it is categorical. 

In Table~\ref{tab:civil-fine} we plot the worst-group performance of different methods when evaluated over these $16 + 2 \cdot {^8\textrm{C}_2} = 72$ groups.  Hence, this evaluation is verifying whether methods that may be robust to group shifts defined by binary attributes, are also robust to shifts when groups are defined by combinations of binary attributes. First, we find that the performance of every method drops, including that of the oracle \gdro which assumes knowledge of the $8$ demographic identities. This observation is in line with some of the findings in prior work~\cite{kearns2018preventing}. Next, similar to Table~\ref{tab:sc-datasets}, we see that the performance of \bdro is still significantly better than CVaR DRO where the adversary is unconstrained. Finally, this experiment provides evidence that the bitrate-constraint does not restrict the adversary from identifying groups defined solely by binary attributes. Our simple group shift assumption (Assumption~\ref{assm:simple-group-shift}) is still satisfied when the spurious attribute is not binary as long as the group $G$ corresponds to an indicator function $\mathbb{I}(\mathbf{x}, y \in G)$ (e.g., an intersection of hyperplanes) that is realized in a low-bitrate class (e.g., a neural net with a VIB constraint which is how we model the adversary in the VIB version of BR-DRO). 

\newcommand{\cc}[1]{\textcolor{black}{#1}}

\begin{table}[!ht]
    \centering
    \footnotesize
    \begin{tabular}{r|c}
        \cc{Method} &  \cc{Worst-group accuracy}  \\
        \cc{ERM} & \cc{52.8  \mt{0.8}} \\
        \cc{LfF ~\citep{nam2020learning}}  & \cc{51.7 \mt{1.0}} \\
        \cc{RWY ~\citep{idrissi2022simple}} & \cc{61.9 \mt{0.8}} \\
        \cc{JTT ~\citep{liu2021just}} & \cc{60.5 \mt{0.9}} \\
        \cc{CVaR DRO ~\citep{levy2020large}} & \cc{56.5 \mt{0.7}} \\ \midrule 
        \cc{\bdro (\vib) (ours)} & \cc{62.9 \mt{0.8}} \\
        \cc{\bdro (\ltwop) (ours)} & \cc{62.5 \mt{0.9}} \\ \midrule
        \cc{\gdro ~\citep{sagawa2019distributionally}} &  \cc{63.0 \mt{0.8}}
    \end{tabular}
    \caption{\textcolor{black}{\textbf{\bdro is robust to group shifts defined by multiple simple attributes:} On the CivilComments test set we evaluate the worst-group accuracy of \bdro and baselines when the groups can comprised of at most two demographic identities and the toxicity label. In ($\cdot$) we report the standard error of the mean accuracy across five runs.}} 
    \label{tab:civil-fine}
\end{table}

\subsection{Comparing \bdro with other baselines that do not assume access to group labels.}
\label{appsubsec:george-bpa}

\cc{
In Section~\ref{sec:experiments} we compare \bdro with baselines JTT~\cite{liu2021just}, RWY~\cite{idrissi2022simple}, LfF~\cite{nam2020learning} and CVaR DRO~\cite{levy2020large} with regards to their performance on datasets that exhibit known spurious correlations (CelebA, Waterbirds and CivilComments) in Table~\ref{tab:sc-datasets} as well as on domain generalization datasets (FMoW and Camelyon17) that present unspecified covariate shifts in Table~\ref{tab:gen-cov-shifts}. None of the above baselines assume access to group annotations on training data. Here, we look at two additional baselines that also do not assume group annotations on training samples: George~\citet{sohoni2020no} and BPA~\citet{seo2022unsupervised}\footnote{\cc{For both baselines, we use the publicly available implementations from the authors of the original works.}}. Both these baselines employ a two-stage method to learn debiased representations where the first stage involves estimating biased pseudo-attributes using a clustering algorithm based on the observation that for a sufficiently trained model, non-target attributes tend to have similar representations. The second stage trains an unbiased model by optimizing a re-weighted objective where weights assigned to each cluster are updated with an exponential moving average, similar to \gdro. BPA also accounts for the size of each cluster when assigning the weights.     
}

\cc{
We evaluate George and BPA on Waterbirds, CelebA and the hybrid dataset FMoW. While these methods were developed with the motivation of tackling group shifts along spurious attributes, following our experiments in Section~\ref{subsec:gen-cov-shift}, we also test how they do on domain generalization kind of tasks. The comparisons with both versions of \bdro are presented in Table~\ref{tab:george-bpa}. On the worst-group accuracy metric, we find that \bdro outperforms George on all datasets and BPA on Waterbirds and FMoW, while being comparable with BPA on CelebA. Since both these methods do not up-weight arbitrary points with high losses, we can think of them as having an implicit constraint on their weighting schemes (adversary), thus yielding solutions that are less pessimistic than CVaR DRO. At the same time, unlike the robust set outlined in our Assumption~\ref{assm:simple-group-shift}, and the excess risk results (in Section~\ref{sec:analysis}), it is unclear what the precise robust sets are for BPA and George, as well as the excess risk of their learned solutions.   
}

\begin{table}[!ht]
    \centering
    \footnotesize
    \begin{tabular}{r|cccccc}
         &  \multicolumn{2}{c}{\cc{Waterbirds}} & \multicolumn{2}{c}{\cc{CelebA}} & \multicolumn{2}{c}{\cc{FMoW}}  \\
        \cc{Method}                             & \cc{Avg}        & \cc{WG} & \cc{Avg}  & \cc{WG} & \cc{Avg} & \cc{WG} \\ \midrule
        \cc{George ~\citep{sohoni2020no}}       & \cc{94.8 \mt{0.4}} & \cc{77.3 \mt{0.6}} & \cc{92.8 \mt{0.3}} & \cc{64.9 \mt{0.7}} & \cc{50.5 \mt{0.4}} & \cc{30.6 \mt{0.5}} \\
        \cc{BPA ~\citep{seo2022unsupervised}}   & \cc{93.7 \mt{0.5}} & \cc{85.2 \mt{0.4}} & \cc{88.0 \mt{0.4}} & \cc{81.7 \mt{0.5}} & \cc{51.3 \mt{0.3}} & \cc{30.7 \mt{0.3}} \\ \midrule
        \cc{\bdro (\vib) (ours)}                & \cc{94.1 \mt{0.2}} & \cc{86.3 \mt{0.3}} & \cc{86.7 \mt{0.2}} & \cc{80.9 \mt{0.4}} & \cc{52.0 \mt{0.2}} & \cc{31.8 \mt{0.2}} \\
        \cc{\bdro (\ltwop) (ours)}              & \cc{93.8 \mt{0.2}} & \cc{86.4 \mt{0.3}} & \cc{87.7 \mt{0.3}} & \cc{80.4 \mt{0.6}} & \cc{53.1 \mt{0.1}} & \cc{32.3 \mt{0.2}}  
    \end{tabular}
    \caption{\textcolor{black}{We compare the performance of baseline methods George~\citep{sohoni2020no} and BPA~\citep{seo2022unsupervised} with \bdro on Waterbirds, CelebA and FMoW. We show both average (Avg) and worst group (WG) accuracy. For FMoW, we show the worst-region (W-Reg) accuracy. In ($\cdot$) we report the standard error of the mean accuracy across five runs.}} 
    \label{tab:george-bpa}
\end{table}

%% file: sections/proofs.tex
\newpage
\vspace{0.2in}
\section{Omitted Proofs}
\label{appsec:omitted-proofs}

First we shall state some a couple of technical lemmas that we shall refer to at multiple points. Then, we prove our theoretical claims in our analysis Section~\ref{sec:analysis}, in the order in which they appear. Before we get into those we provide proof for our Corollary~\ref{cor:group-struc-unique} and the derivation of Bitrate-Constrained CVaR DRO in Equation~\ref{eq:bitcon-cvar-dro}.



\vspace{0.2in}

\begin{lemma}[Hoeffding bound~\cite{wainwright2019high}]
\label{lem:hoeffding}
Let $X_1, \ldots, X_n$ be a set of $\mu_i$ centered independent sub-Gaussians,  each with parameter $\sigma_i$. Then for all $t\geq 0$, we have
\begin{align}
    \Prob \brck{\frac{1}{n}\sum_{i=1}^{n} (X_i - \mu_i) \geq t} \leq \exp\paren{-\frac{n^2t^2}{2\sum_{i=1}^{n}\sigma_i^2}}.
\end{align}
\end{lemma}

\vspace{0.2in}

\begin{lemma}[Lipschitz functions of Gaussians~\cite{wainwright2019high}]
\label{lem:functions-of-gaussian}
Let $X_1, \ldots, X_n$ be a vector of \iid Gaussian variables and $f:\Real^n \mapsto \Real$ be $L$-Lipschitz with respect to the Euclidean norm. Then the random variable $f(X) - \E[f(X)]$ is sub-Gaussian with parameter at most $L$, thus:
\begin{align}
    \Prob[|f(X) - \E[f(X)]| \geq t] \leq 2\cdot \exp{\paren{-\frac{t^2}{2L^2}}}, \;\; \forall \, t\geq 0.
\end{align}
\end{lemma}
\vspace{0.2in}

\subsection{Proof of Corollary~\ref{cor:group-struc-unique}}

Let us recall the definition of a well defined group structure. For a pair of measures $Q \ll P$ we say $\gG(P, Q)$ is well defined if given there exists a set of disjoint measurable sets $\gG_{P, Q} = \{G_k\}_{k=1}^{K}$ such that $G_k \in \Sigma$, $Q(G_k)>0$, $Q(\gG(P, Q))=1$ and we have:

\begin{align}
    K = \min \{ |\{G_1, \ldots, G_M\}| \st  p(\rvx,\ry \mid G_m) = q(\rvx,\ry \mid G_m) > 0, \forall (\rvx, \ry) \in G_m\;\; \forall m \in [M] \}
\end{align}

Now by definition $K$ is finite. Thus if there exists two well defined group structures $\gG_1(P, Q)$ and $\gG_2(P, Q)$ for the same pair $P, Q$ then  it must be the case that $K = \gG_1(P, Q)$ = $\gG_2(P, Q)$. 

Then, there must exist $G \in \gG_1(P, Q)$ such that $Q(G) > 0$ and $G', G'' \in \gG_2(P, Q)$ where $Q(G'), Q(G'') >0$ and $Q(G \cap G'), Q(G \cap G'') > 0$.

Note that since $G, G', G'' \in \Sigma$ that is closed under countable unions, we have that $G \cap G'$ and $G \cap G''$ are two sets where $q(\rvx, \ry) >0$ $\forall (\rvx, \ry) \in G\cap G', G \cap G''$.

Let $(\rvx_1, \ry_1) \in (G \cap G')$ and $(\rvx_2, \ry_2) \in (G \cap G'')$. From definition we know that $q(\rvx_2, \ry_2), q(\rvx_1, \ry_1)>0$ and . Since both $(\rvx_1, \ry_1)$ and $(\rvx_2, \ry_2)$ are in $G$ we have that:

\begin{align}
 q(\rvx_1, \ry_1)=\frac{Q(G)}{P(G)} \cdot p(\rvx_1, \ry_1) = \frac{Q(G')}{P(G')} \cdot p(\rvx_1, \ry_1)\\
 q(\rvx_2, \ry_2)=\frac{Q(G)}{P(G)} \cdot p(\rvx_2, \ry_2) = \frac{Q(G'')}{P(G'')} \cdot p(\rvx_2, \ry_2)
\end{align}

Thus, we can conclude that $\frac{Q(G')}{P(G')} = \frac{Q(G'')}{P(G'')}$. This implies that $G' \cup G''$ also satisfies the following that $Q(G' \cup G'') > 0$ and $q(\rvx, \ry \mid G' \cup G'') = p(\rvx, \ry \mid G' \cup G'')$. 

Thus, we can construct a new $\gG_3(P,Q) = \{G \in \gG_2(P, Q) \st G\notin \{G', G''\}\} \cup \{G' \cup G''\}$. Clearly, $\gG_3(P, Q)$ satisfies all group structure properties and is smaller than $\gG_2(P, Q)$. Thus, we arrive at a contradiction which proves the claim that $\gG(P,Q)$ is indeed unique whenever well defined.

\vspace{0.3in}
\subsection{Derivation of Bitrate-Constrained CVaR DRO in \eqref{eq:bitcon-cvar-dro}}

Recall that we define $\gW$ as the set of all measurable functions $w \st \gX \times \gY \mapsto [0,1]$, since the other convex restrictions in \eqref{eq:adv-constraints} are handled by dual variable $\eta$. As in Section~\ref{sec:bdro}, $\gW(\gamma)$ is derived from the new $\gW$ using Definition~\ref{def:bitrate-constrained-class}. With that let us first state the CVaR objective~\citep{levy2020large}.

\begin{align}
   & \gL_{\textrm{cvar}}(h, P) \coloneqq \sup_{q} \int_{\gX \times \gY} q(\rvx, \ry) \cdot l(h) \nonumber \\
   & \textrm{s.t.} \;\;  q \geq 0,\;\; \|q/p\|_\infty \leq (1/\alpha_0), \;\; \int_{\gX \times \gY} q(\rvx, \ry)  = 1
\end{align}

The objective in $q$ is linear with convex constraints, and has a strong dual (see ~\citet{duchi2016statistics,boyd2004convex} for the derivation) which is given by:

\begin{align}
     & \quad  \inf_{\eta \in \Real} \cbrck{ \frac{1}{\alpha_0} \E_P (l(h) - \eta)_+ + \eta } \nonumber  \\
    &= \inf_{\eta \in \Real} \cbrck{ \frac{1}{\alpha_0} \innerprod{(l(h) - \eta)_+}{\I}_P + \eta } \nonumber  \\
    &= \inf_{\eta \in \Real} \cbrck{ \frac{1}{\alpha_0} \innerprod{(l(h) - \eta)}{\I(l(h) - \eta \geq 0)}_P + \eta } \\
    &= \inf_{\eta \in \Real} \sup_{w \in \gW} \cbrck{ \frac{1}{\alpha_0} \innerprod{(l(h) - \eta)}{w}_P + \eta } 
\end{align}

The last equality is true since the set $\I(l(h) - \eta \geq 0)$ is measurable under $P$ (based on our setup in Section~\ref{sec:prelim}). Note that for any $h$, the objective $\frac{1}{\alpha_0} \innerprod{(l(h) - \eta)}{w}_P + \eta$ is linear in $w$, and $\eta$. If we further assume the loss $l(h)$ to be the $\lzone$ loss, it is bounded, and thus the optimization over $\eta$ can be restricted to a compact set. Next, $\gW$ is also a compact set of functions since we restrict our solvers to measurable functions that take values bounded in $[0, 1]$. 
\begin{align}
    \gL_{\textrm{cvar}}(h, P) = \inf_{\eta \in \Real} \sup_{w \in \gW} \cbrck{ \frac{1}{\alpha_0} \innerprod{(l(h) - \eta)}{w}_P + \eta } 
\end{align}
The above objective is precisely the Bitrate-Constrained CVaR DRO objective we have in \eqref{eq:bitcon-cvar-dro}. Later in the Appendix we shall need an equivalent form of the objective which we shall derive below. 

We can now invoke the Weierstrass' theorem in \citet{boyd2004convex} to give us the following:

\begin{align}
    \gL_{\textrm{cvar}}(h, P) = \inf_{\eta \in \Real} \sup_{w \in \gW} \cbrck{ \frac{1}{\alpha_0} \innerprod{(l(h) - \eta)}{w}_P + \eta }  \nonumber \\
    =  \frac{1}{\alpha_0} \sup_{w \in \gW} \cbrck{  \inf_{\eta \in \Real} \innerprod{(l(h) - \eta)}{w}_P + \eta }    
\end{align}

Now,  the final objective $\inf_{h\in \gH}  \gL_{\textrm{cvar}}(h, P)$ is given by:

\begin{align}
   \frac{1}{\alpha_0}  \inf_{h\in \gH} \sup_{w \in \gW} \cbrck{  \inf_{\eta \in \Real} \innerprod{(l(h) - \eta)}{w}_P + \eta }    
\end{align}

In the above equation we can now replace the unconstrained class $\gW$ with our bitrate-constrained class $\wgam$ to get the following:

\begin{align}
   \frac{1}{\alpha_0}  \inf_{h\in \gH} \sup_{w \in \wgam} \cbrck{  \inf_{\eta \in \Real} \innerprod{(l(h) - \eta)}{w}_P + \eta }    
\end{align}

\vspace{0.3in}
\subsection{Proof of Theorem~\ref{thm:worst-risk-gen}}
\label{appsubsec:omitted-proofs-5.1}

For convenience we shall first restate the Theorem here.

\begin{theorem}[[restated] worst-case risk generalization]
With probability $\geq 1-\delta$ over sample $\gD \sim P^n$, the worst risk for $\hD^\gamma$ can be upper bounded by the following oracle inequality:
{\footnotesize
\begin{align}
     \sup_{w \in \dwgam} R(\hD^\gamma, \etaD, w) \;-\; \gL^*_{\textrm{cvar}}(\gamma) \;\lsim \;  \frac{M}{\alpha_0}  \sqrt{\paren{\gamma + \log\paren{\frac{1}{\delta}} + (d+1) \log\paren{\frac{L^2n}{\gamma}} +  \log n}/{(2n -1)}} \nonumber,
\end{align}}
when $l(\cdot, \cdot)$ is $[0,M]$-bounded, $L$-Lipschitz and $\gH$ is parameterized by convex set $\Theta \subset \Real^d$.
    \label{thm:restated-worst-risk-gen}
\end{theorem}

The overview of the proof can be split into two parts:
\begin{itemize}
    \item For each learner, first obtain the oracle PAC-Bayes~\citep{mcallester1998some} worst risk generalization guarantee over the adversary's action space $\dwgam$.
    \item Then, apply uniform convergence bounds using a union bound over a covering of the class $\gH$ to get the final result.
\end{itemize}

\textbf{Intuition:} The only tricky part lies in the fact that oracle PAC-Bayes inequality would not give us arbitrary control over the generalization error for each learner, which we would typically get in Hoeffding type bounds. Hence, we need to ensure that the the worst risk generalization rate decays faster than how the size of the covering would increase for a ball of radius defined by the worst generalization error. 

Now, we shall invoke the following PAC-Bayes generalization guarantee stated (Lemma~\ref{lem:pac-bayes}) since $R(h, \eta, w) \in [0, M/\alpha_0]$.

\vspace{0.2in}
\begin{lemma}[PAC-Bayes~\citep{catoni2007pac,mcallester1998some}] 
    \label{lem:pac-bayes}
        With probability $\geq 1-\delta$ over choice of dataset $\gD$ of size $n$ the following inequality is satisfied
        \begin{align}
            \E_P\E_Q(\lzone(h(\rvx), \ry)) \leq \E_{\pn}\E_Q(\lzone(h(\rvx), \ry)) + \sqrt{\frac{D(Q||P) + \log (1/\delta) + \frac{5}{2} \log n +8}{2n-1}}
        \end{align}
\end{lemma}
\vspace{0.2in}

A direct application of this gives us that with probability at least $1-\omega$:
.
\begin{align}
    &\E_{w \sim \delta} R(h, \eta, w)   \leq \E_{w \sim \delta} \brck{ \frac{1}{\alpha_0}\innerprod{l(h)-\eta}{w}_\pn } + \eta + \sqrt{\frac{\kl{\delta}{\pi} + \log (1/\omega) + \frac{5}{2} \log n +8}{2n-1}} \nonumber
\end{align}

Let $\hat{R}_D(h, \eta, w) = \frac{1}{\alpha_0}\innerprod{l(h)-\eta}{w}_\pn  + \eta  $
Since the above inequality holds for any data dependent $\delta$:. 
\begin{align}
    & \sup_{\delta \in \dwgam}\E_{w \sim \delta} R(h, \eta, w) \leq \sup_{\delta \in \dwgam} \brck{ \hat{R}_D(h, \eta, w) + \eta + \sqrt{\frac{\kl{\delta}{\pi} + \log (1/\omega) + \frac{5}{2} \log n +8}{2n-1}}} \nonumber
\end{align}

Further, we make use of the fact $\kl{\delta}{\pi} \leq \gamma$.
\begin{align}
    \leq \sup_{\delta_1 \in \dwgam} \brck{ \hat{R}_D(h, \eta, w)} + \sup_{\delta_2 \in \dwgam} \brck{ \sqrt{\frac{\kl{\delta_2}{\pi} + \log (1/\omega) + \frac{5}{2} \log n +8}{2n-1}}} \nonumber 
\end{align}

Thus,
\begin{align}
    \sup_{\delta \in \dwgam}\E_{w \sim \delta} R(h, \eta, w) - \sup_{\delta \in \dwgam}\E_{w \sim \delta} \hat{R}_D(h, \eta, w) \leq  \brck{ \sqrt{\frac{\gamma + \log (1/\delta) + \frac{5}{2} \log n +8}{2n-1}}} \nonumber 
\end{align}

To actually apply this uniformly over $h, \eta$, we would first need two sided concentration which we derive below as follows:

Let $a_i = \hat{R}_D(h, \eta, \delta) -  R(h, \eta, \delta)$, Since $R(h, \eta, \delta) \leq M/\alpha_0$, we can apply Hoeffding bound with $t = \lambda/n$ in Lemma~\ref{lem:hoeffding} on $a_i$ to get: 

\begin{align}
    \E_\gD{\exp{(\lambda\cdot a_i)}} \leq \exp{\frac{\lambda^2 (M/\alpha_0)^2}{8n}} 
   \E_\pi \E_\gD{\exp{(\lambda\cdot a_i)}} \leq \E_\pi \exp{\frac{\lambda^2 (M/\alpha_0)^2}{8n}} 
\end{align}

Applying Fubini's Theorem, followed by the Donsker Varadhan variational formulation we get:

\begin{align}
 \E_\gD \E_\pi \brck{\exp{(\lambda\cdot a_i)}} \leq \E_\pi \exp{\frac{\lambda^2 (M/\alpha_0)^2}{8n}} \\
 = \E_\gD \exp{\sup_{\delta \in \dwgam} \brck{{(\lambda\cdot a_i)} - \kl{\delta}{\pi}}} \leq  \exp{\frac{\lambda^2 (M/\alpha_0)^2}{8n}} 
\end{align}

The Chernoff bound finally gives us with probability $\geq 1- \omega$:

\begin{align}
    \E_{\pn}\E_Q((h(\rvx), \ry)) \lsim  \E_P\E_Q((h(\rvx), \ry)) +  \frac{M}{\alpha_0} \sqrt{\frac{\kl{\delta}{\pi} + \log (1/\omega) + \log n}{2n-1}}
\end{align}

Using the reverse form of the empirical PAC Bayes inequality, we can do a derivation similar to the one following the PAC-Bayes bound in Lemma~\ref{lem:pac-bayes} to get for any fixed $\eta \in [0,M], h \in \gH$ we get:

\begin{align}
    & \abs{\sup_{\delta \in \dwgam}\E_{w \sim \delta} R(h, \eta, w) - \sup_{\delta \in \dwgam}\E_{w \sim \delta} \hat{R}_D(h, \eta, w)} \lsim \frac{M}{\alpha_0} \sqrt{\frac{\kl{\delta}{\pi} + \log (1/\omega) + \log n}{2n-1}} \\
    & \fourquad \lsim  \frac{M}{\alpha_0} \sqrt{\frac{\gamma + \log (1/\omega) + \log n}{2n-1}}
\end{align}

Because we see that in the above bound the dependence on $\delta$, is given by a $\log$ term we are essentially getting an "exponential-like" concentration. So we can think about applying uniform convergence bounds over the class $\gH \times [0, M]$ to bounds the above with high probability $\forall (h, \eta)$ pairs. 

We will now try to get uniform convergence bounds with two approaches that make different assumptions on the class of functions $l(h)$. The first is very generic and we will show why such a generic assumption is not sufficient to get an upper bound on the generalization that is $\gO(1/\sqrt{n})$ in the worst case. Then, in the second approach we show how assuming a parameterization will fetch us a rate of that form if we additionally assume that the loss function is $L$-Lipschitz.

\underline{Approach 1:} 

Assume $l(h)$ lies in a class of $(\alpha, 1)$-H\"{o}lder continuous functions 
Now we shall use the following covering number bound for  $(\alpha, 1)$-H\"{o}lder continuous functions to get a uniform convergence bound over $\gH \times [0,M]$. 

\begin{lemma}[Covering number $(\alpha, 1)$-H\"{o}lder continuous] 
    \label{lem:convering-number}
        Let $\gX$ be a bounded convex subset of $\Real^d$ with non-empty interior. Then, there exists a constant $K$ depending only on $\alpha$ and $d$ such that 
        \begin{align}
            \log \gN(\epsilon, C_1^\alpha(\gX), \|\cdot\|_\infty) \leq K \lambda(\gX^1) \paren{\frac{1}{\epsilon}}^{d/\alpha} 
        \end{align}
        for every $\epsilon > 0$, where $\lambda(\gX^1)$ is the Lebesgue measure of the set $\{x \st \|x - \gX\| \leq 1\}$. Here, $C_1^\alpha(\gX)$ refers to the class of $(\alpha, 1)$-H\"{o}lder continuous functions.
\end{lemma}
\vspace{0.2in}

We assume that $l(h)$ is $(\alpha, 1)$-H\"{o}lder continuous. And therefore by definition, of $R(h, \eta, \cdot)$, the function is $(\alpha, 1)$-H\"{o}lder continuous in $(l(h), \eta)$. Similat argument applies for $\sup_{\delta \in \dwgam}\E_{w \sim \delta} R(h, \eta, w)$ since taking a pointwise supremum for a linear function over a convex set $\Delta(\gW, \gamma)$ would retain H\"{o}lder continuity for some value of $\alpha$. Applying the above we get: 

\begin{align}
   \log \gN(\epsilon, \sup_{\delta \in \dwgam}\E_{w \sim \delta} R(\cdot, \cdot, w), \|\cdot\|_\infty) \lsim \paren{\frac{M}{\alpha_0} \sqrt{\frac{\gamma + \log (1/\omega) + \log n}{2n-1}}}^{-(d/\alpha)}
\end{align}

Now, we can show that with probability at least $1-\delta$, $\forall h \in \gH$ we get:

\begin{align}
    &\abs{\sup_{\delta \in \dwgam}\E_{w \sim \delta} R(h, \eta, w) - \sup_{\delta \in \dwgam}\E_{w \sim \delta} \hat{R}_D(h, \eta, w)} \\
     &\fourquad \lsim \frac{M}{\alpha_0} \sqrt{\frac{\gamma + \log (  \gN(\epsilon, R(\cdot, \cdot, w), \|\cdot\|_\infty)/\delta) + \log n}{2n-1}} \\
       &\fourquad \lsim \frac{M}{\alpha_0} \sqrt{\frac{\gamma + \paren{\paren{\frac{M}{\alpha_0} \sqrt{\frac{\gamma + \log (1/\delta) + \log n}{2n-1}}}^{-(d/\alpha)}} + \log (1/\delta) + \log n}{2n-1}}
\end{align}

Note that in the above bound we cannot see if this upper bound shrinks as $n \rightarrow \infty$, without assuming something very strong about $\alpha$. Thus, we need covering number bounds that do not grow exponentially with the input dimension. And for this we turn to parameterized classes, which is the next approach we take. It is more for the convenience of analysis that we introduce the following parameterization. 

\underline{Approach 2:} 

Let $l(\cdot, \cdot)$ be a $[0, M]$ bounded $L$-Lipschitz function in $\|\cdot\|_2$ over $\Theta$ where $\gH$ be parameterized by a convex subset $\Theta \subset \Real^d$. Thus we need to get a covering of the loss function $\sup_\delta \E_{w \sim \delta} R(\theta, \eta, w)$ in $\|\cdot\|_\infty$ norm, for a radius $\epsilon$. A standard practice is to bound this with a covering $\gN(\Theta, \frac{\epsilon}{L}, \|\cdot\|_2)$, where $\|\cdot\|_2$ is Euclidean norm defined on $\Theta \subset \Real^d$. 

\begin{lemma}[Covering number for $\gN(\Theta \times \brck{0, M}, \frac{\epsilon}{L}, \|\cdot\|_2)$~\cite{wainwright2019high}] 
    \label{lem:convering-number-2}
        Let $\Theta$ be a bounded convex subset of $\Real^d$ with . 
        \begin{align}
            \gN(\epsilon/L, \Theta, \|\cdot\|) \lsim \paren{1+\frac{L}{\epsilon}}^{d+1} 
        \end{align}
\end{lemma}
\vspace{0.2in}

We now re-iterate the steps we took previously:

\begin{align}
    &\abs{\sup_{\delta \in \dwgam}\E_{w \sim \delta} R(h, \eta, w) - \sup_{\delta \in \dwgam}\E_{w \sim \delta} \hat{R}_D(h, \eta, w)} \\
     &\fourquad \lsim \frac{M}{\alpha_0} \sqrt{\frac{\gamma + \log (  \gN(\epsilon, R(\cdot, \cdot, w), \|\cdot\|_\infty)/\delta) + \log n}{2n-1}} \\
       &\fourquad \lsim \frac{M}{\alpha_0} \sqrt{\frac{\gamma + \log \paren{1+\frac{L}{\sqrt{\gamma/n}}}^{d+1} + \log (1/\delta) + \log n}{2n-1}} \\
       &\fourquad \lsim \frac{M}{\alpha_0} \sqrt{\frac{\gamma + (d+1) \log \paren{\frac{L^2n}{\gamma}} + \log (1/\delta) + \log n}{2n-1}}
\end{align}

Note that the above holds with probability atleast $1-\delta$ and for $\forall h, \eta$. Thus, we can apply it twice:

\begin{align}
  &  \abs{\sup_{\delta \in \dwgam}\E_{w \sim \delta} R(\hD^\gamma, \etaD, w) - \sup_{\delta \in \dwgam}\E_{w \sim \delta} \hat{R}_DR(\hD^\gamma, \etaD, w)} \\
  &   \fourquad  \lsim  \frac{M}{\alpha_0} \sqrt{\frac{\gamma + (d+1) \log \paren{\frac{L^2n}{\gamma}} + \log (1/\delta) + \log n}{2n-1}}
\end{align}
\begin{align}
   & \abs{\sup_{\delta \in \dwgam}\E_{w \sim \delta} R(h^*, \eta^*, w) - \sup_{\delta \in \dwgam}\E_{w \sim \delta} \hat{R}_DR(h^*, \eta^*, w)} \\
   & \fourquad \lsim  \frac{M}{\alpha_0} \sqrt{\frac{\gamma + (d+1) \log \paren{\frac{L^2n}{\gamma}} + \log (1/\delta) + \log n}{2n-1}}
\end{align}

where $h^*, \eta^*$ are the optimal for $\gL_{\textrm{cvar}}^*$. Combining the two above proves the statement in Theorem~\ref{thm:worst-risk-gen}.

\vspace{0.4in}

\subsection{Proof of Theorem~\ref{thm:special-case-rkhs}}
\label{appsubsec:omitted-proofs-5.2}


\textbf{Setup.} Let us assume there exists a prior $\Pi$ such that $\wgam$ in Definition~\ref{def:bitrate-constrained-class} is given by an RKHS induced by  Mercer kernel $k:\gX\times\gX \mapsto\Real$, s.t. the eigenvalues of the kernel operator decay polynomially: 
\begin{align}
\mu_j \lsim j^{{-2}/{\gamma}}    
\end{align}
 for $(\gamma < 2)$. We solve for $\hD^\gamma,\etaD$ by doing kernel ridge regression over norm bounded  ($\|f\|_{\wgam}$$\leq$$M$) smooth functions $f$. Thus, $\wgam$ is compact.

\begin{align}
    & \argmax_{w \in \gW(\gamma)), \|w\|_{\wgam}\leq R}  \;\;\; R(h, \eta, w) \;\; = \;\; \argmax_{w \in \gW(\gamma)), \|w\|_{\wgam}\leq R}   \;\;\;  \innerprod{l(h)-\eta}{w}_P + \eta \\
    & \qquad \argmax_{w \in \gW(\gamma), \|w\|_{\wgam}\leq R}  \;\;\; \E_P \I((l(h)-\eta) \cdot w > 0)
    \label{eq:krr-obj-class}
\end{align}
 
 We show that we can control: (i) the pessimism of the learned solution; and (ii) the generalization error (Theorem~\ref{thm:special-case-rkhs}). 
 Formally, we refer to pessimism for estimates $\hD^\gamma, \etaD$:
 \begin{align}
     \textrm{excess risk or pessimism:} \;\;\; \sup_{w \in \wgam}|\inf_{h,\eta} R(h, \eta, w) - R(\hD^\gamma, \etaD, w)|
 \end{align}

\begin{theorem}[(restated for convenience) bounded RKHS]
    \label{thm:res-special-case-rkhs} For $l, \gH$ in Theorem~\ref{thm:worst-risk-gen}, and for $\wgam$ described above $\exists$ $\gamma_0$ s.t. for all sufficiently bitrate-constrained $\gW(\gamma)$ \ie $\gamma$$\leq$$\gamma_0$, w.h.p.  $1- \delta$ worst risk generalization error is 
    $\gO\paren{(1/n)\paren{\log(1/\delta) + (d+1) \log(nR^{-\gamma} L^{\gamma/2})}}$ 
    and the excess risk is $\gO(M)$ for $\hD^\gamma, \etaD$ above. 
\end{theorem}

\vspace{0.2in}
\underline{Generalization error proof:}

Note that the objective in \eqref{eq:krr-obj-class} is a non-parametric classification problem. We can convert this to the following non-parametric regression problem, after replacing the expectation with plug-in $\pn$. 

\begin{align}
    \inf_{w \in \wgam), \|w\|_{\wgam}\leq R} \frac{1}{n} \sum_{i=1}^{n} (w(x_i, y_i) - (l(h(x_i), y_i) - \eta) + \epsilon_i)^2 + \lambda_n \|w\|^2_{\wgam}
    \label{eq:krr-obj-reg}
\end{align}

where $\lambda_n$ $\rightarrow 0$ as $n \rightarrow \infty$. Essentially, for non-parametric kernel ridge regression regression the regularization can be controlled to scale with the critical radius, that would give us better estimates and tighter localization bounds as we will see. 

Note that in the above problem we add variable $\epsilon_i$ which represents random noise $\sim \gN(0, \sigma_2)$. Let $\sigma_2=1$ for convenience. Since the noise is zero mean and random, any estimator maximizing the above objective on $\pn$ would be consistent with the estimator that has a noise free version. We can also thing of this as a form regularization (similar to $\lambda$), if we consider the kernel ridge regression problem as the means to obtain the Bayesian predictive posterior under a Bayesian prior that is a Gaussian Process $\gG\gP(\bf{0}, \sigma_2 k(x, x))$, under the same kernel as defined above. 

First we will show estimation error bounds for the following KRR estimate:

\begin{align}
    \hat{w}_D^\gamma  = \argmin_{w \in \wgam), \|w\|_{\wgam}\leq R} \frac{1}{n} \sum_{i=1}^{n} (w(x_i, y_i) - (l(h(x_i), y_i) -\eta) + \epsilon_i)^2 + \lambda_n \|w\|^2_{\wgam}
    \label{eq:krr-obj-reg-2}
\end{align}

The estimation error would be measured in terms of $\pn$ norm \ie $\|\hat{w}_D^\gamma - w^*\|_\pn$ where 
\begin{align}
{w_*}^\gamma(x, y) = \argmin_{w \in \wgam), \|w\|_{\wgam}\leq R} \;\; \E_P \E_{\epsilon} ((l(h(x), y)-\eta)-w(x,y) + \epsilon)^2    
\end{align}
is the best solution to the optimization objective in population.

\textbf{Next steps:}
\begin{itemize}
    \item First, we get the estimation error in $\|\hat{w}_D^\gamma - {w_*}^\gamma\|_\pn$  of $\pn$.
    \item Then using uniform laws ~\citep{wainwright2019high} we can extend it to $L^2(P)$ norm \ie $\|\hat{w}_D^\gamma - w^*\|_p$.
    \item Then we shall prove that if we convert the $\hat{w}_D^\gamma$ and $w^*$ into prediction rules: $\hat{w}_D^\gamma \geq 0$ and ${w_*}^\gamma$, then we can get the estimation error of prdedictor $\hat{w}_D^\gamma \geq 0$ with respect to the optimal decision rule ${w_*}^\gamma \geq 0$  in class $\wgam$.
    \item The final step would give us an oracle inequality of the form in Theorem~\ref{thm:worst-risk-gen}.
\end{itemize}

Based on the outline above, let us start with getting $\|\hat{w}_D^\gamma - w^*\|_\pn$. For this we shall use concentration inequalities from localization bounds (see Lemma~\ref{lem:loc-conc}).
Before we use that, we define the quantity $\delta_n$, which is the critical radius (see Ch. 13.4 in ~\cite{wainwright2019high}). For convenience, we also state it here. Formally, $\delta_n$ is the smallest value of $\delta$ that satisfies the following inequality (critical condition):
\begin{align}
\label{eq:crit-con}
    \frac{\gR_n(\delta)}{\delta} \leq \frac{R}{2} \cdot \delta
\end{align}
where,
\begin{align}
    \gR_n(\delta) \coloneqq \E_{\epsilon} \brck{\sup_{g \in (\gF - f^*), \|g\|_\gF \leq R,\;
    \|g\|_\pn \leq \delta} \abs{\frac{1}{n} \sum_{i=1}^n \epsilon_i g(x_i, y_i) \cdot l(h(x_i) - y_i)}} 
\end{align}
and $\epsilon$ is some sub-Gaussian zero mean random variable.

\begin{lemma}[~\citep{wainwright2019high}]
\label{lem:loc-conc} For some convex RKHS class $\gF$
Let $\hat{f}$ be defined as:
\begin{align}
    \hat{f} \in \argmin_{f\in \gF, \|f\|_\gF\leq R} \cbrck{\frac{1}{n} \sum_{i=1}^n (y_i - f(x_i))^2 + \lambda_n \|f\|^2_\gF}
\end{align}
then, with probability $\geq 1- c_2 \exp\paren{{-c_3} \frac{nR^2\delta_n^2}{\sigma^2}}$ and when $\lambda_n \geq \delta_n^2$
we get: 
\begin{align}
    \|\hat f - f^*\|_2^2 \leq c_0 \inf_{\|\gF\|\leq R} \|f - f^*\|_n^2 + c_1 R^2 (\delta_n^2 + \lambda_n).
\end{align}
\end{lemma}

Note that it is standard exercise in statistics to derive the following closed form for the problem in \eqref{eq:krr-obj-reg-2}:

\begin{align}
    \hat{w}_D^\gamma(\cdot) = \hat{K}_n(\cdot, Z) (\hat{K}_n^T\hat{K}_n + \lambda_n {I})^{-1} \paren{l(h)_D-\epsilon_D}
\end{align}
where $l(h)_D$ is the loss vector and $\epsilon_D$ is the noise vector for dataset $\gD$ and $\hat{K}_n$ is the empirical kernel matrix given by $\hat{K}_{i,j} = \frac{1}{n} k((x_i, y_i), (x_j, y_j))$, and $Z$ is a matrix of $(x, y)$ pairs in dataset. 

\begin{corollary}{~\cite{wainwright2019high}} 
    Let $\hat \mu_j$ be the eigen values $\hat{\mu}_1 \geq \hat{\mu}_2  \ldots \geq \hat{\mu}_n$ for the empirical Kernel matrix $\hat{K}$, then we have for any $\delta$ satisfying
    \begin{align}    
    \sqrt{\frac{2}{n} \paren{\sum_{i=1}^n \min(\delta^2, \hat{u}_j) }} \;\; \leq \;\; \frac{R}{4} \delta^2
    \end{align},
    it is necessary that $\delta$ satisfies the critical condition in \eqref{eq:crit-con}. 
\end{corollary}

To show the above critical condition we shall now use the polynomial decaying property that for the specific kernel induced by $\wgam$, as stated in our assumption in the beginning of this section. For this we take standard approach taken for polynomial decay kernels ~\cite{zhang2013divide}. Let $\exists C$ for some large $C > 0$ such that $\hat{{\mu}}_j \leq C j^{-2/\gamma}$. Then for some $k$, such that $\delta^2 \geq c k^{-2/\gamma}$ 

\begin{align}
    & \sqrt{\frac{1}{n} \paren{\sum_{j=1}^n \min(\delta^2, \hat{\mu}_j) }} \quad \lsim \quad \sqrt{\frac{2}{n} \paren{\sum_{i=1}^n \min(\delta^2, C j^{-2/\gamma}) }}  \\
    & \qquad \lsim \sqrt{\frac{2}{n} \paren{ k\delta^2 + C \sum_{j=k+1}^n  j^{-2/\gamma}) }} \quad \lsim \quad \sqrt{\frac{2}{n} \paren{ k\delta^2 + C \sum_{j=k+1}^{\infty}  j^{-2/\gamma}) }} \\
    & \qquad \lsim \sqrt{\frac{2}{n} \paren{ k\delta^2 + C \int_{j=k+1}^{\infty}  z^{-2/\gamma}\; dz) }} \quad \lsim \quad  \sqrt{\frac{2}{n} \paren{ k\delta^2 + C  k^{-2/\gamma + 1}\; dz) }} \\ 
    & \qquad \leq \sqrt{2/n} \paren{ \sqrt{k} \cdot {\delta}} \leq \frac{1}{\sqrt{n}} \cdot \delta^{1-\gamma/2}
\end{align}

Now, setting the above into the critical condition equation from Corollary above:
\begin{align}
   & \frac{1}{\sqrt{n}} \cdot \delta^{1-\gamma/2} \leq \frac{R}{4} \delta^2 \\
    \qquad &\implies  \delta^{1+\gamma/2} \geq \frac{1}{\sqrt{n}R} 
\end{align}

This tells us that: \begin{align}
    \label{eq:crit-rad}
    \delta_n^2 \gsim \paren{\frac{1}{nR^2}^{\frac{2}{\gamma+2}}}
\end{align} is the critical radius. 

We shall later plug this into the bound we have into a uniform bound over the concentration inequality in Lemma~\ref{lem:loc-conc}. 
The reason we need a uniform bound over Lemma~\ref{lem:loc-conc} is that in its current form, it only bounds  $\|\hat{w}^\gamma_D - w_*^\gamma\|^2_\pn$ for a specific choice of $\eta, h$. In order to arrive at the worst risk generalization error of the form we have in Theorem~\ref{thm:worst-risk-gen} we need to satisfy that with high probability $1-\delta$ $\forall \eta, h$, a critical concentration bound of the form in Lemma~\ref{lem:loc-conc} but over $\sup_{\eta, h} \|\hat{w}^\gamma_D - w_*^\gamma\|^2_\pn $. 

Let $\epsilon = c_2 \exp\paren{c_3 nR^2 \frac{{\delta_n}^2}{\sigma^2}}$. Since $\delta_n^2$ needs to be large enough (see condition in \eqref{eq:crit-rad}), we use Lemma~\ref{lem:loc-conc} in the following bound, incorporating $\delta_n$ condition we derived. 

With high probability $1-\epsilon$:
\begin{align}
      \|\hat{w}^\gamma_D - w_*^\gamma\|^2_\pn \lsim \inf_{w \in \wgam, \|w\|\leq R} \|w - w^\gamma_*\|_\pn^2 +  R^2  \max \paren{\paren{\frac{1}{nR^2}}^{\frac{\gamma+2}{\gamma}}, \paren{\log(1/\epsilon)\frac{1}{nR^2}}} 
    \label{eq:bnd}
\end{align} 

To apply uniform convergence argument on the above we would need to apply a union bound on a covering of $\Theta \times \brck{0,M}$, so that we get the probability bound to hold for all $\eta, h$. 

For this we use the same technique as in the proof of Theorem~\ref{thm:worst-risk-gen}. First, we shall use Lemma~\ref{lem:convering-number-2} to get a covering number bound for bounded convex subset $\Theta$ of $\Real^d$ that parameterizes the learner (Theorem~\ref{thm:special-case-rkhs}) . 
        
\begin{align}
    \gN(\beta/L, \Theta \times [0, M], \|\cdot\|) \lsim \paren{1+\frac{L}{\beta}}^{d+1} 
\end{align}

And we know that a covering of $\Theta \times \brck{0,M}$ in radius $\beta / L$, will fetch a covering for $l(h)-\eta$ in $\beta$, since we assume $l(\cdot)$ to be Lipschitz in $\theta$. Thus, all we need to prove bound \eqref{eq:bnd} holds uniformly is to get a covering in radius $ R^2  \max \paren{\paren{\frac{1}{nR^2}}^{\frac{2}{\gamma+2}}, \paren{\log(1/\epsilon)\frac{1}{nR^2}}} $. Thus, a acovering in $R^2 \paren{\paren{\frac{1}{nR^2}}^{\frac{2}{\gamma+2}}}$. Thus, the number of elements in cover are:

\begin{align}
   J =  \paren{1+\frac{L}{\paren{R^2 \paren{\frac{1}{nR^2}^{\frac{2}{\gamma+2}}}}}}^{d+1}
\end{align}

For union bound we need:

\begin{align}
 &   J  \epsilon/c_2  = \exp\paren{-c_3 nR^2\delta_n^2}  \\
  &  \implies \log (\frac{1}{\epsilon}) + \log J \gsim c_3 nR^2\delta_n^2 \\
  &  \implies  \log (\frac{1}{\epsilon}) + (d+1) \log \paren{\frac{L}{\paren{R^2 \paren{\paren{\frac{1}{nR^2}}^{\frac{2}{2+\gamma}}}}}} \gsim c_3 nR^2\delta_n^2 \\
    &  \implies  \log (\frac{1}{\epsilon}) + (d+1) \log \paren{{\paren{LR^{-2}}^{\frac{\gamma+2}{2}} nR^2}} \gsim c_3 nR^2\delta_n^2 
\end{align}

The uniform convergence bound that we get is $ R^2  \max \paren{\paren{\frac{1}{nR^2}}^{\frac{\gamma+2}{\gamma}}, \paren{\log(J/\epsilon)\frac{1}{nR^2}}}$. In the above sequence of steps we have shown that, due to the size of $J$, the second term would be maximum, or at least there exists  a $\gamma_0$, such that the second term would be higher for all $\gamma \geq \gamma_0$, for any sample size. 

Thus, we get the following probabilistic uniform convergence. With probability $\geq 1-\epsilon$, $\forall \eta, h:$

\begin{align}
 & \|\hat{w}^\gamma_D - w_*^\gamma\|^2_\pn \lsim \inf_{w \in \wgam, \|w\|\leq R} \|w - w^\gamma_*\|_\pn^2 \\ 
 & \qquad \lsim    \frac{1}{n} \log (\frac{1}{\epsilon}) + (d+1) \log \paren{{\paren{LR^{-2}}^{\frac{\gamma+2}{2}} nR^2}}   \\
 & \qquad \lsim    \frac{1}{n} \log (\frac{1}{\epsilon}) + (d+1) \log \paren{{\paren{L^{\gamma/2} R^{-\gamma}} n}}   
\end{align}

Applying the above twice, once one $\hat{w}_D^\gamma$ and another on ${w}_*^\gamma$ we prove the generalization bound in Theorem~\ref{thm:special-case-rkhs}. 

\vspace{0.2in}
\underline{Excess risk bound:}

In the same setting we shall now prove the excess risk bound. Recall the definition of excess risk:

\begin{align}
\textrm{excess risk} \coloneqq \sup_{w \in \wgam}|\inf_{h,\eta} R(h, \eta, w) - R(\hD^\gamma, \etaD, w)|.     
\end{align}

Let $h^*(w), \eta^*(w) = \inf_{h,\eta} R(h, \eta, w)$, then:

\begin{align}
    \textrm{excess risk} &= \sup_{w \in \wgam}|\inf_{h,\eta} R(h, \eta, w) - R(\hD^\gamma, \etaD, w)| \\
  \qquad   &\leq \sup_{w \in \wgam} \paren{R(h^*(w), \eta^*(w), w) - R(\hD^\gamma, \etaD, w)}  \\
  \qquad   &\leq \sup_{w \in \wgam} \paren{\frac{1}{\alpha_0} \innerprod{l(h^*) - l(\hD^\gamma) - (\eta^*(w) - \etaD)}{w}_P} \\
  \qquad   &\leq \frac{M}{\alpha_0} \sup_{w \in \wgam} \paren{(\|w\|_{L_2(P)})}
    \label{eq:rk-2}
\end{align}

Note, that according to our assumption $\|w\|_{\wgam} \leq B$ \ie the smooth functions are bounded in RKHS norm. The following lemma relates bounds in RKHS norm to bound in $L_2(P)$ bound for kernels with bounded operator norms:

\begin{lemma}
    \label{lem:rk-3}
    For an RKHS $\gH_k$ with norm $\|\cdot\|_{\gH_k}$:
    \begin{align}
        \|f\|_{L^2(P)} = \|T_K^{1/2} f\|_{\gH_k} \leq \sqrt{\|T_K^{1/2}\|_{\textrm{op}}} \cdot \|f\|_{\gH_k}
    \end{align}
\end{lemma}

\textit{Proof:}
\begin{align}
    &\|T_K^{1/2} f\|_{\gH_k}^2 = \innerprod{T_K^{1/2} f}{T_K^{1/2} f}_{\gH_k} =\innerprod{f}{T_K f}_{\gH_k} \\
    & = \sum_{j=1}^{\infty} \frac{\innerprod{\phi_j}{f}_{L^2(P)}, \innerprod{\phi_j}{T_K f}_{L^2(P)}} {\lambda_j} \\
    & = \|f\|_{L^2(P)}^2
\end{align}
In the above $\lambda_j$ are the Eigen values of the kernel and the Eigen functions $\phi_j$ are orthonormal and span $L^2(P)$/
Thus, $\|f\|_{L^2(P)} \leq \|T_K^{1/2}\|_{op} \cdot \|f\|_{\gH_k}$. Since we assume polynomially decaying Eigen values for our kernel, it is easy to see that $\|T_K^{1/2}\|_{op} = \gO(1)$. 

Applying Lemma~\ref{lem:rk-3} to \eqref{eq:rk-2}, directly gives us the excess risk bound and completes the proof.

\begin{align}
    \textrm{excess risk} \lsim  \|T_K^{1/2}\|_{op} \cdot B  =  \gO(B)
\end{align}


\vspace{0.3in}
\subsection{Proof of Lemma~\ref{lem:nash-eq}}

First let us re-state the optimization objective which the two players: learner $h \in \gH$ and adversary $w \in \wgam$ are trying to optimize via an online game (Section~\ref{sec:analysis}). Next, we use the fact that $R(h, \eta, w)$ is linear in $w$. 

\begin{align}
     \gL^*_{\textrm{cvar}}(\gamma)  &=  \inf_{h \in \gH, \eta \in \Real}  \sup_{w \in \gW(\gamma)}  R(h, \eta, w) 
    =  \inf_{h \in \gH, \eta \in \Real}  \sup_{w \in \dwgam}  R(h, \eta, w)\;\; \nonumber \\
    &  \qquad \textrm{where,}\;\;
    R(h, \eta, w) = (1/\alpha_0) \innerprod{l(h)-\eta}{w}_P + \eta
\end{align} 

Hence, the adversary is learning a mixed strategy (probabilistic). Now, recall that $l(\cdot)$ is a strictly convex objective over the space of learners $\gH$. Under the negative entropy regularizer, the objective above has a unique saddle point since $\gH$ and $\wgam$ are compact sets (\citet{rockafellar1970convex}). This saddle point is also exactly the Nash equilibrium of the two player game.

\vspace{0.3in}
\subsection{Proof of Theorem~\ref{thm:convergence-excess-guarantee}}
\label{appsubsec:omitted-proofs-5.4}

\textbf{Setup.} The algorithm is as follows: Consider a two-player zero-sum game where the learner uses a no-regret strategy to first play $h \in \gH, \eta \in \Real$ to minimize $\E_{w \sim \delta} R(h, \eta, w)$. Then, the adversary plays follow the regularized leader (FTRL) strategy to pick distribution $\delta \in \dwgam$ to maximize the same. The regularizer used is a negative entropy regularizer.
Our goal is to analyze the bitrate-constraint $\gamma$'s effect on the above algorithm's convergence rate and the pessimistic nature of the solution found. For this, we need to first characterize the bitrate-constraint class $\gW(\gamma)$. 
So we assume there exists a prior $\Pi$ such that $\wgam$ is Vapnik-Chervenokis (VC) class of dimension $O(\gamma)$.

Note that $R(h,\eta,w)$ is convex in $h$ and linear in $\eta, l$. Thus, as we discuss in the derivation for \eqref{eq:bitcon-cvar-dro} this objective optimized over convex sets has a unique saddle point (Nash equilibrium) by Weierstrass's theorem. Thus, to avoid repetition we only discuss the proofs for the other two claims on convergence and excess risk.

\underline{Convergence:}

Given that $\wgam$ is a VC class of dimension $C\gamma$ for some large $C$, we can use Sauer-Shelah~\cite{bartlett1997covering} Lemma (stated) below to bound the total number of groups that can be identified by $\wgam$ in $n$ points.

\begin{lemma}[Sauer's Lemma]
    \label{lem:sauer}
    The Vapnik-Chervonenkis dimension of a class $\gF$, denoted as VC-dim($\gF$), and it is the cardinality of the largest set $S$ shattered by $\gF$. Let $d = VC-dim(\gF)$, then for all $m$, $C[m] = \gO(m^d)$
\end{lemma}

Thus, the total number of groups that can be proposed on $n$ points by $\wgam$ is ${\gO}(n^\gamma)$. A similar observation was made in \citet{kearns2018preventing}. Different from them, our goal is to analyze the algorithm iterates for our solver described above and bound its pessimism. 

First, for convergence rate we show that the above algorithm has a low regret---a standard exercise in online convex optimization.
Note that any distribution picked by the adversary can be seen as multinomial over a finite set of possible groups that is let's say $K$, and from discussion above we know that $K=O(n^\gamma)$. Further, the negative entropy regularizer is given as:

\begin{align}
    B(\delta) \coloneqq c \cdot \sum_{i=1}^{K} \delta_i \log \delta_i
\end{align}

where the sum is over total possible groups identified by $\wgam$. Let the probability assigned to group $i$ be denoted as $\delta_i$. The FTRL strategy for adversary is given as:

\begin{align}
    \delta_T = \argmin_{\delta \in \Delta(\wgam)} \sum_{t=1}^{T-1} \frac{1}{\alpha_0} \innerprod{l(h_t)-\eta_t}{ \delta_t}_{\pn} + \eta + c \cdot \sum_{i=1}^{K} \delta_i \log \delta_i
\end{align}

Then the regret for not having picked a single action $\delta$ is given as:

\begin{align}
    \textrm{REGRET}_T(\delta) \coloneqq \sum_{t=1}^{T} \frac{1}{\alpha_0} \innerprod{l(h_t)-\eta_t}{\delta_t - \delta_{t+1}}_\pn +  B(\delta) - B(\delta_1)
\end{align}

We bound the two terms in the above bound separately. With $\sum_{k=1}^{k} \delta_k = 1$, we get the strong dual for the FTRL update above as:

\begin{align}
   \sum_{t=1}^{T-1} \frac{1}{\alpha_0} \innerprod{l(h_t)-\eta_t}{ \delta_t}_{\pn} + \eta + c \cdot \sum_{i=1}^{K} \delta_i \log \delta_i
+ \lambda \cdot (\sum_{i=1}^{K} \delta_i -1)
\end{align}

Solving we get:

\begin{align}
    \delta_t(k) = \frac{\exp\paren{\frac{-1}{c}} \sum_{t=1}^{t-1} \E_\pn \frac{1}{\alpha_0}(l(h_t) - \eta_t | G_k) + \eta/K}{\sum_{k=1}^{K}\exp\paren{\frac{1}{\alpha_0}\frac{-1}{c}} \sum_{k=1}^{t-1} (\E_\pn \frac{1}{\alpha_0} (l(h_t) - \eta_t | G_k) + \eta/K)}
\end{align}

where $\E_\pn (l(h_t) - \eta_t | G_k)$ is the expected empirical loss in group $G_k$ and $\delta_t(k)$ is the adversary's distribution at time step $t$ for the $k^{th}$ group.

Claim on stability:

\begin{align}
    \frac{1}{\alpha_0} \innerprod{l(h_t)-\eta_t}{\delta_t - \delta_{t+1}}_\pn  \leq 1 / c
\end{align}

The above statement is true because, 
\begin{align}
\delta_{t+1}(i)= \delta_t(i) \cdot \exp\paren{\frac{1}{\alpha_0 c} \E[l(h_t)-\eta_t | G_i] + \eta_t/K}    
\end{align}

Thus, if $l(h_t) \in [0,M/\alpha_0]$, \ie losses are bounded then:

\begin{align}
    \delta_{t+1}(i) \geq \delta_{t}(i) \cdot e^{-1/c}  \geq \delta_{t}(i) \cdot (1-1/c). 
\end{align}

and our stability claim is easy to see. Thus, we have bounded the first term in our regret bound above. Further, we can to see that $B(x) - B(x_1) \leq c \log K $. Thus, we have bounded both terms in the regret bound above in terms of $c$.

\begin{align}
    \textrm{REGRET}_T \leq (T/c) + (c \log K)
\end{align}

Setting  $c = \sqrt{\frac{T}{\log K}}$, we get:

\begin{align}
    \frac{\textrm{REGRET}_T}{T} \leq \sqrt{\frac{\log K}{T}}
\end{align}

Now, our VC claim gave $K = \gO(n^\gamma)$. Hence,

\begin{align}
    \frac{\textrm{REGRET}_T}{T} = \gO \sqrt{\frac{\gamma \log n}{T}}
\end{align}

Next, we use Theorem 9 from \citet{abernethy2018faster} that maps low regret $O(\epsilon)$ algorithms in zero-sum convex-concave games to $\epsilon$-optimal equilibriums.

Let regret be $\epsilon$, then applying their theorem gives us:

\begin{align}
    V^*  - \epsilon \leq \inf_{h \in \gH, \eta \in \Real} R_D(h, \eta, \bar{\delta}_{T}) \leq V* \leq \sup_{\delta \in \Delta(\wgam)} R_D(\bar{h}_T, \bar{\eta}_T, \delta) \leq V^* + \epsilon
\end{align}

where 
\begin{align}
    V^* = R_D(h^*_D(\gamma), \eta^*_D(\gamma), \delta^*_D(\gamma)) =  \inf_{h \in \gH, \eta \in \Real} \sup_{\delta \in \Delta(\wgam)} \frac{1}{\alpha_0} \innerprod{l(h)-\eta}{\delta} + \eta
\end{align}

\vspace{0.2in}
\underline{Excess risk:}

For excess risk we need to bound:

\begin{align}
    & \frac{1}{\alpha_0} \sup_{h \in \gH, \eta \in \Real} \abs{ \sup_{\delta \in \Delta(\wgam)} \innerprod{l(h)-\eta}{\delta - {\delta}^*(\gamma}| } \\
    & \qquad \leq \frac{M}{\alpha_0} \frac{1}{2}\textrm{TV}(\delta - {\delta}^*(\gamma)) \leq \frac{M}{2 \alpha_0}(1-1/K)
    = \frac{M}{ \alpha_0} \gO(1-1/n^\gamma)
\end{align}

In the above argument we used the fact that at equilibrium, ${\delta}^*(\gamma)$ would be uniform over all possible distinct group assignments. 
This completes our proof of Theorem~\ref{thm:convergence-excess-guarantee}.

\subsection{Worst-case generalization risk for Group DRO}

Recall that the \bdro objective in \eqref{eq:bdro-main} involves an expectation over $P$, which in practice is replaced by empirical distribution $\hat{P}_n$. This induces errors in estimating the worst-case risk for plug-in estimates $\hD^\gamma, \etaD$  
In Theorem~\ref{thm:worst-risk-gen} we saw how the bitrate-constraint gracefully controls the worst-case generalization guarantee for estimates $\hD^\gamma, \etaD$ through an oracle inequality. Here, we ask: ``How does the worst-case generalization for Group DRO compare with the bound in Theorem~\ref{thm:worst-risk-gen}?''.

First, let us recall the objective for \gdro (\eqref{eq:gdro-main}) which assumes the knowledge of ground-truth groups $G_1, G_2, \ldots, G_K$.
\begin{align}
    \gL^*_{\mathrm{gdro}}\; \coloneqq \; \inf_{h \in \gH} \sup_{k \in [K]} \E_P \brck{l(h(\rvx), \ry) \mid (\rvx, \ry) \in G_k}
    \label{eq:gdro-main}
\end{align}
Now, let us denote the plug-in estimate as $\hD^K$ which solves the above objective using the empirical distribution $\hat{P}_n$. We are now ready to state Theorem~\ref{thm:worst-risk-gen-gdro} which gives us the worst-risk generalization error for \gdro. 
\begin{theorem}[worst-case risk generalization (Group DRO)]
With probability $\geq 1-\delta$ over $\gD \sim P^n$, the worst group risk for $\hD^K$ can be upper bounded by the following oracle inequality:
{\footnotesize
\begin{align}
     \sup_{k \in [K]} \E_P \brck{l(\hD^K(\rvx), \ry) \mid (\rvx, \ry) \in G_k} \;\lsim \; \gL^*_{\mathrm{gdro}} +  M  \sqrt{\paren{\log\paren{\frac{2K}{\delta}} + (d+1) \log\paren{1+L^2n} }/{(2n)}}\; \nonumber,
\end{align}}
when $l(\cdot, \cdot)$ is $[0,M]$-bounded, $L$-Lipschitz and $\gH$ is parameterized by convex set $\Theta \subset \Real^d$.
    \label{thm:worst-risk-gen-gdro}
\end{theorem}

\textit{Proof Sketch.}

We can use the covering number result in Lemma~\ref{lem:convering-number}, and plug it into the Hoeffding bound in Lemma~\ref{lem:hoeffding} to bound the generalization gap for a specific group $G_k$. This yields the following result with probability at least $1 - \delta$:
\begin{align}
    & \abs{\E_P \brck{l(h(\rvx), \ry) \mid (\rvx, \ry) \in G_k} - \E_{\hat{P}_n} \brck{l(h(\rvx), \ry) \mid (\rvx, \ry) \in G_k}} \\
    & \fourquad \lsim \sqrt{\frac{1}{2n} \cdot \paren{ \log \paren{\frac{2}{\delta}} + (d+1) \log \paren{ {1+L^2n}}}}
\end{align}

Next, we use union bound over the $K$ groups to get the final result in Theorem~\ref{thm:worst-risk-gen-gdro}, with probability at least $1-\delta$. 

The main difference between the generalization analysis \bdro and \gdro is the order in which we bound the errors for the learner and the adversary (groups). For \gdro, we begin with the learner since we can use traditional uniform convergence bounds to bound per group generalization risk and then use union bound over the $K$ groups to bound the worst-case risk. We can do this precisely since we can use the pre-determined groups $G_1, G_2, \ldots, G_K$ assumed by \gdro. On the other hand, since we do not assume this knowledge for \bdro, we first use PAC-Bayes bound (Lemma~\ref{lem:pac-bayes})  to control the generalization error per group through the bitrate constraint $\gamma$. The PAC-Bayes bound allows us to reason about the generalization error for a specific learner $h \in \gH$. Finally, to get the bound in Thorem~\ref{thm:worst-risk-gen} we relied on a covering number argument for $\gH$ parameterized as a convex subset of $\Real^d$ (Section ~\ref{appsubsec:omitted-proofs-5.1}).